%% file: sample-sigconf.tex
\documentclass[sigconf, screen]{acmart}

\usepackage{lipsum}
\usepackage{caption}
\usepackage{subcaption}
\usepackage{enumitem}
\usepackage{bm}
\usepackage{xcolor}
\usepackage{booktabs}
\usepackage{multirow}
\usepackage{arydshln}
\usepackage{pifont}

\newcommand{\ie}{\textit{i}.\textit{e}.}
\newcommand{\eg}{\textit{e}.\textit{g}.}
\newcommand{\etc}{\textit{e}.\textit{t}.\textit{c}.}

\usepackage{array} 
\newcolumntype{x}[1]{>{\centering\arraybackslash}p{#1pt}}

\AtBeginDocument{%
  \providecommand\BibTeX{{%
    \normalfont B\kern-0.5em{\scshape i\kern-0.25em b}\kern-0.8em\TeX}}}

\setcopyright{acmcopyright}
\setcopyright{acmlicensed}
\setcopyright{rightsretained}
\copyrightyear{2022}
\acmYear{2022}
\acmDOI{https://doi.org/10.1145/3503161.3547756}

\copyrightyear{2022}
\acmYear{2022}
\setcopyright{acmcopyright}
\acmConference[MM '22] {Proceedings of the 30th ACM International Conference on Multimedia }{October 10--14, 2022}{Lisbon, Portugal.}
\acmBooktitle{Proceedings of the 30th ACM International Conference on Multimedia (MM '22), October 10--14, 2022, Lisbon, Portugal}
\acmPrice{15.00}
\acmISBN{978-1-4503-9203-7/22/10}
\acmDOI{10.1145/3503161.3547756}
\acmSubmissionID{66}

\begin{document}


\title{Correspondence Matters for Video Referring Expression Comprehension}

\author{Meng Cao$^1$, Ji Jiang$^1$, Long Chen$^{{2}}$, Yuexian Zou$^{{1,3}}$}
\affiliation{%
\institution{$^1$ School of Electronic and Computer Engineering, Peking University, $^2$Columbia University, $^3$Peng Cheng Laboratory}
\city{}
\country{}
}
\email{{mengcao,zouyx}@pku.edu.cn, jiangji@stu.pku.edu.cn, zjuchenlong@gmail.com}

\renewcommand{\shortauthors}{Trovato and Tobin, et al.}


\input{sec/0_abs}




\maketitle

\input{sec/1_intro}

\input{sec/2_related}
\input{sec/3_method}
\input{sec/4_exp}
\input{sec/5_con}
\input{sec/6_appendix}

\clearpage
\balance
\bibliographystyle{ACM-Reference-Format}
\bibliography{acmart}

\end{document}

%% file: sec/0_abs.tex
\begin{abstract}
We investigate the problem of video Referring Expression Comprehension (REC), which aims to localize the referent objects described in the sentence to visual regions in the video frames. Despite the recent progress, existing methods suffer from two problems: 1) inconsistent localization results across video frames; 2) confusion between the referent and contextual objects. To this end, we propose a novel Dual Correspondence Network (dubbed as \textbf{DCNet}) which explicitly enhances the dense associations in both the inter-frame and cross-modal manners. Firstly, we aim to build the inter-frame correlations for all existing instances within the frames. Specifically, we compute the inter-frame patch-wise cosine similarity to estimate the dense alignment and then perform the inter-frame contrastive learning to map them close in feature space. Secondly, we propose to build the fine-grained patch-word alignment to associate each patch with certain words. Due to the lack of this kind of detailed annotations, we also predict the patch-word correspondence through the cosine similarity. Extensive experiments demonstrate that our DCNet achieves state-of-the-art performance on both video and image REC benchmarks. Furthermore, we conduct comprehensive ablation studies and thorough analyses to explore the optimal model designs. Notably, our inter-frame and cross-modal contrastive losses are plug-and-play functions and are applicable to any video REC architectures. For example, by building on top of Co-grounding~\cite{song2021co}, we boost the performance by 1.48\% absolute improvement on Accu.@0.5 for VID-Sentence dataset. Our codes are available at \url{https://github.com/mengcaopku/DCNet}. 
\end{abstract}

%% file: sec/1_intro.tex
\section{Introduction}
\input{table_figs/figTeaser}

Referring Expression Comprehension (REC)~\cite{hu2017modeling,hu2016natural,yu2016modeling,yu2017joint} is a challenging task which aims to localize the region of the image described by the natural language (\ie, referent). It has been extensively studied in both industry and academia, due to its wide applications in downstream tasks, \eg, visual question answering~\cite{antol2015vqa,chen2019counterfactual,chen2022rethinking}, captioning~\cite{chen2017sca,chen2021human}, navigation~\cite{chen2019touchdown}, autonomous driving~\cite{kim2019grounding}, \etc. Recently, several works~\cite{zhou2018weakly,vasudevan2018object,chen2019weakly,zhang2020does} are developed to conduct REC in the video domain. Compared to image REC, video REC is more arduous due to the complex temporal structures.

Although great improvement has been achieved by current methods, it is worth noting that current methods have two overlooked drawbacks: 1) \emph{Inconsistent localization results across video frames.} Compared to image REC, some frames in video REC may suffer from the deteriorated visual quality, \eg, motion blur, light change, object occlusion, and out of focus, leading to unstable and interrupted localization results. Thus, it is hard and error-prone to identify the referent by only utilizing information within one isolated image. A state-of-the-art video REC method \cite{song2021co} also points out this issue and partially alleviates it by introducing the frame-wise attention mechanism. However, such attention fails to build the explicit correspondence and still leads to the sub-optimal performance. For example in Figure.~\ref{fig:teaser1}, the referent ``\emph{elephant}" is obscured in frame \#2 and the method~\cite{song2021co} mistakes another elephant as the object-of-interest. 2) \emph{Confusion between the referent and contextual objects.} Recall that our goal is to ground the object related to the global sentence meaning. It is common for the sentence to contain multiple objects including the \emph{referent} and \emph{contextual objects}. Current methods tend to confuse these two types of objects since they do not introduce detailed alignment between all the mentioned object instances and object areas. As the example shown in Figure.~\ref{fig:teaser2}, the given query sentence (``\emph{A rabbit is running around a deer.}") contains two instance, (\ie, ``\emph{rabbit}", ``\emph{deer}"). The state-of-the-art method~\cite{song2021co} fails to identify the referred instance ``\emph{rabbit}" and falsely localize the contextual object ``\emph{deer}".

\input{table_figs/figSparseDense}

\input{table_figs/figTeaserVis}
Based on these observations, we propose a \textbf{D}ual \textbf{C}orrespondence \textbf{Net}work (dubbed as \textbf{DCNet}) for more accurate video referring expression comprehension. Specifically, an inter-frame contrastive loss is designed to enhance patch-wise\footnote{We use ``\emph{patch}" to denote each spatial feature on the embedded feature maps.} correspondence within frames and a cross-modal contrastive loss is built to model fine-grained word-patch alignment.

Firstly, we contend that \emph{instance feature consistency} is crucial to generate more stable and consistent localization results. Concretely, each instance within the frame should be aware of its potential spatial localization in the contextual frames. Let's revisit the case in Figure.~\ref{fig:teaser1}. Although being obscured, the feature response for the ``\emph{elephant}" in frame \#2 should be consistent with that in frame \#1 and frame \#3. Therefore, we propose an inter-frame contrastive loss to enhance instance correspondence between frames. To achieve this, a straightforward way is to consider the annotated ground-truth areas of each frame as the positive samples (cf. Figure.~\ref{fig:sparseDense} left). This method, however, can not build the \emph{dense} correspondence for each instance. Besides, relying on annotations makes it only applicable to the supervised training scenario. To alleviate this, we take a rather simple manner to estimate the dense correspondence by computing the inter-frame patch-wise cosine similarity. Then the patch pairs with top similarity scores are selected as the positive samples, where their embedding features are mapped close.

Secondly, we believe that building \emph{patch-word} fine-grained alignment helps distinguish the referent and the contextual objects. For example in Figure.~\ref{fig:teaser2}, the query sentence contains two keywords ``\emph{rabbit}" and ``\emph{deer}", which are semantically associated with certain patches in the video frame. Therefore, we propose to model such fine-grained cross-modal alignment. To establish this detailed regularization, we also resort to the cosine similarity computation manner to mine the desired correspondence. Specifically, given one patch feature within the frame, we compute its cosine similarity with each word feature and select those with the highest similarities as positive pairs.

Equipped with the two mentioned contrastive losses, our DCNet constructs both the inter-frame and cross-modal correspondence with the absence of explicit annotations. These two losses could serve as plug-and-play functions that can be easily inserted into existing video REC methods. As the example shown in Figure.~\ref{fig:teaserVis}, by building on top of the baseline model \cite{song2021co}, our inter-frame and cross-modal contrastive losses lead to more consistent and accurate feature responses.

In summary, we make three contributions in this paper:

\begin{itemize}[topsep=0pt, partopsep=0pt, leftmargin=13pt, parsep=0pt, itemsep=3pt]
	\item We contend that the inter-frame and cross-modal alignments are crucial for accurate video REC. To this end, we propose a simple yet effective cosine similarity mining method to extract dense patch-wise and fine-grained patch-word correspondence.

	\item Based on the extracted correspondence, we propose the inter-frame and cross-modality contrastive loss, which boost more consistent video feature representations and optimize fine-grained visual-text relations, respectively.	
	
	\item Experimental results show that our DCNet achieves state-of-the-art performance on both video and image REC benchmarks. Besides, our proposed cross-modal and inter-frame contrastive loss can be further seamlessly integrated into any video REC architectures and achieve promising performance gains.	
\end{itemize}

%% file: table_figs/figTeaser.tex
\begin{figure}[!t]
	\centering
	\begin{subfigure}[b]{0.47\textwidth}
		\centering
		\includegraphics[width=\textwidth]{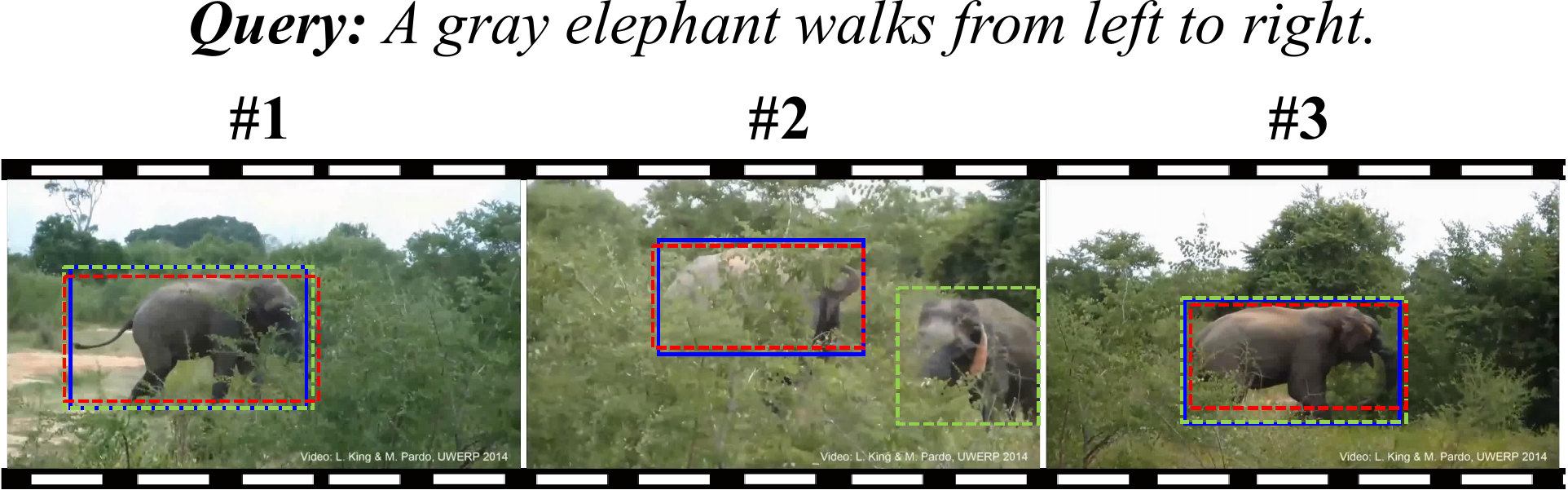}
		\caption{}
		\label{fig:teaser1}
	\end{subfigure}
	\begin{subfigure}[b]{0.47\textwidth}
		\centering
		\includegraphics[width=\textwidth]{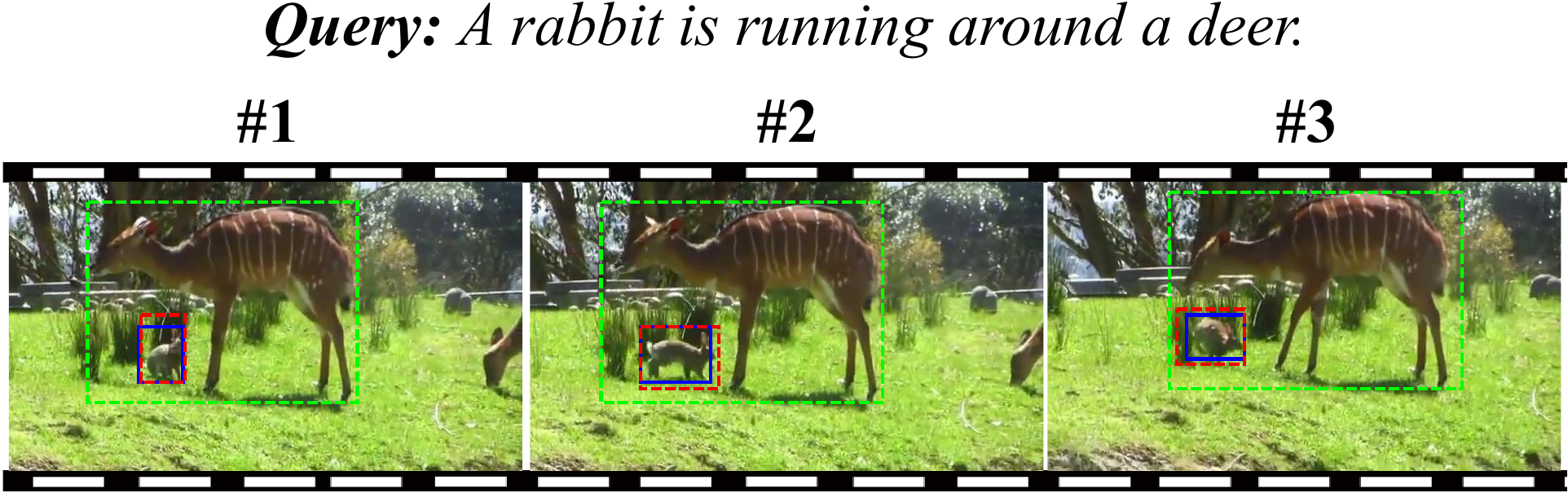}
		\caption{}
		\label{fig:teaser2}
	\end{subfigure}
	\vspace{-2mm}
	\caption{Video REC results of a state-of-the-art model~\cite{song2021co} (marked in \textcolor{green}{green} dotted box) and our DCNet (marked in \textcolor{red}{red} dotted box). The ground-truth annotations are marked in \textcolor{blue}{blue} box. Current SOTA methods have overlooked two issues: (a) Inconsistent localization results across video frames: \cite{song2021co} falsely localizes object-of-interest in frame \#2 due to the occlusion; (b) Confusion between the referent and contextual objects. \cite{song2021co} fails to identify the referred instance ``\emph{rabbit}" and falsely localizes ``\emph{deer}".}
\end{figure}

%% file: table_figs/figSparseDense.tex
\begin{figure}[t]
	\centering
	\includegraphics[width=0.95\columnwidth]{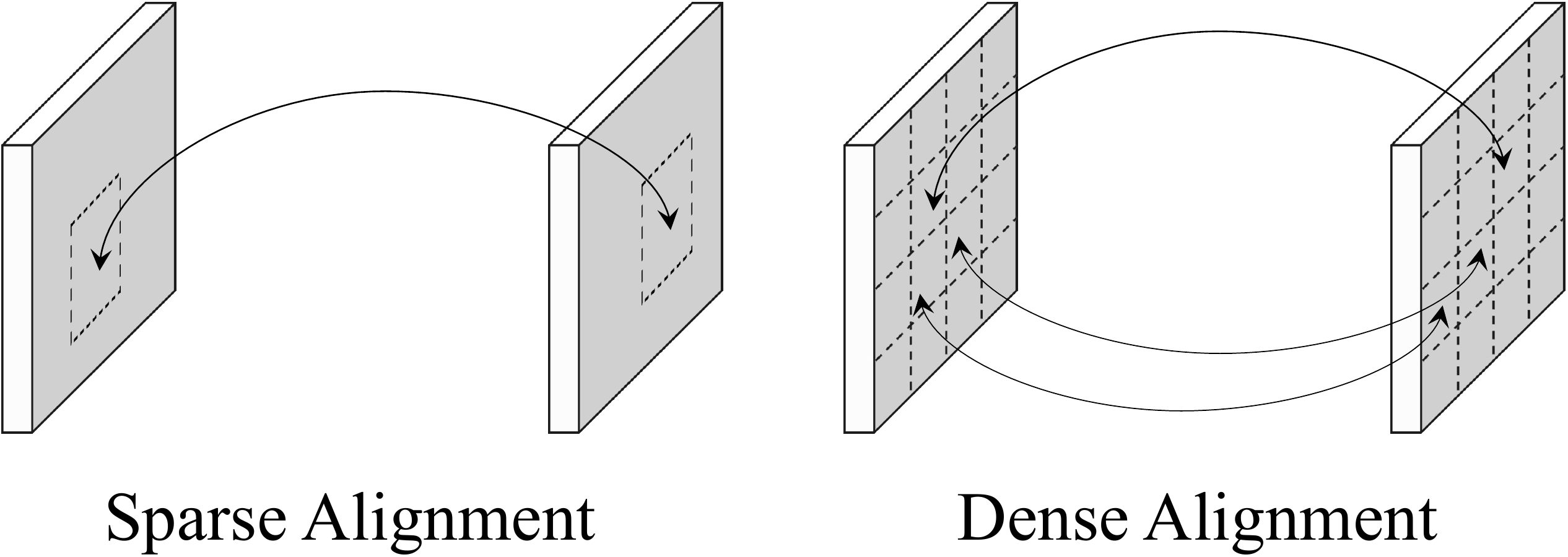}
	\caption{Left: Sparse Alignment is conducted only between the annotated bounding box region features; Right: Dense Alignment is conducted for all the patch features.}
	\label{fig:sparseDense}
\end{figure}

%% file: table_figs/figTeaserVis.tex
\begin{figure}[t]
	\centering
	\begin{subfigure}[b]{0.47\textwidth}
		\centering
		\includegraphics[width=\textwidth]{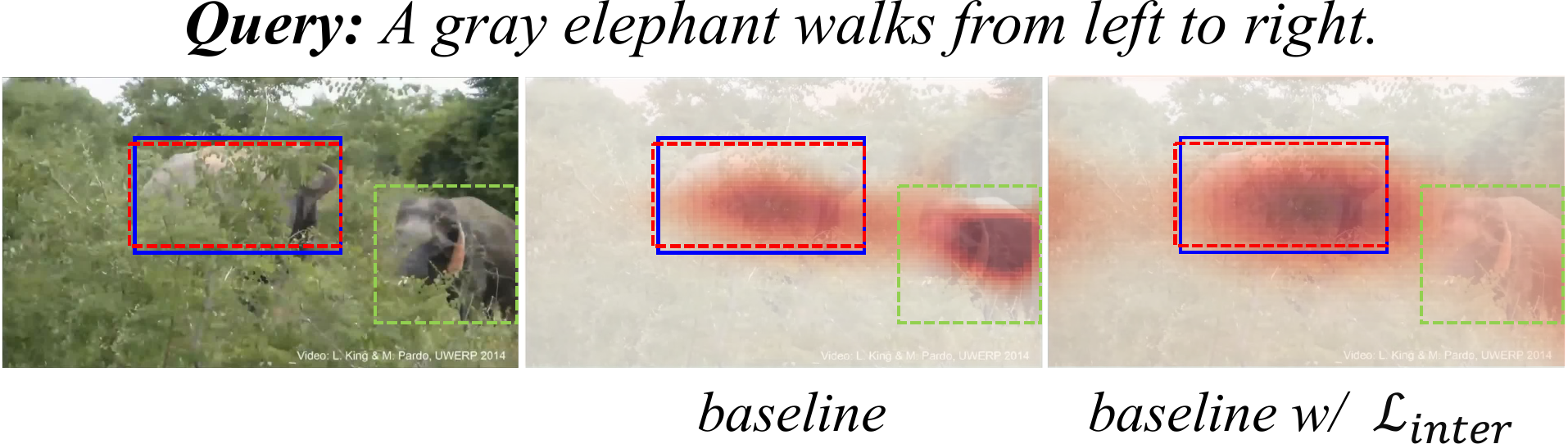}
		\caption{}
		\label{fig:teaserVis1}
	\end{subfigure}
	\begin{subfigure}[b]{0.47\textwidth}
		\centering
		\includegraphics[width=\textwidth]{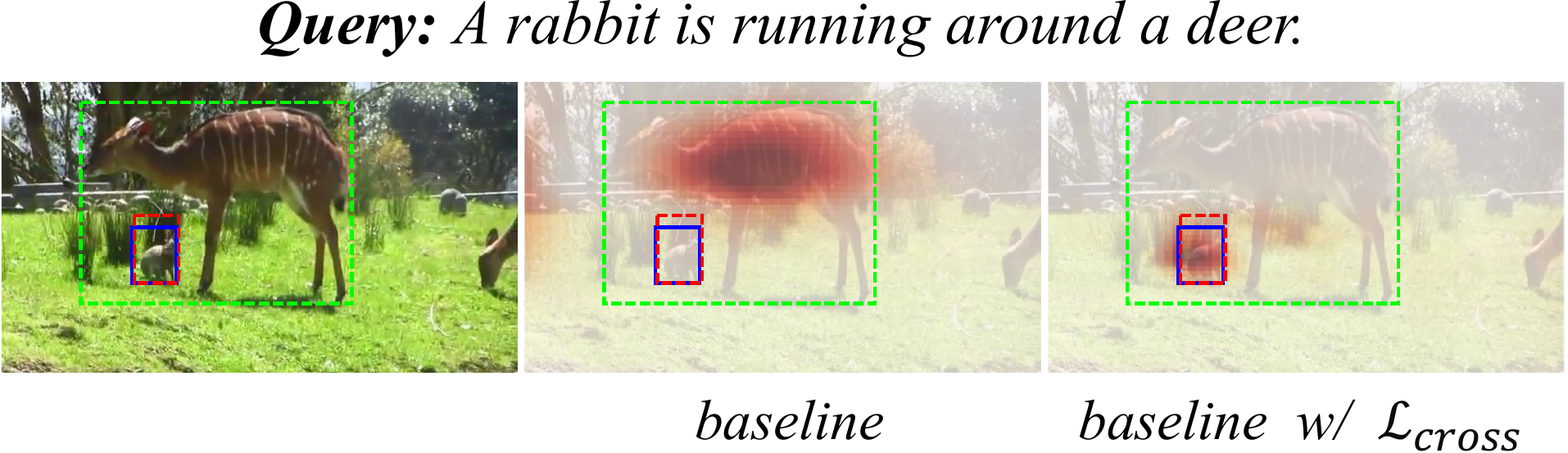}
		\caption{}
		\label{fig:teaserVis2}
	\end{subfigure}
	\vspace{-2mm}
	\caption{Visualizations of confidence score maps of the baseline modal~\cite{song2021co} with (a) our proposed inter-frame contrastive loss $\mathcal{L}_{inter}$ or with (b) cross-modal contrastive loss $\mathcal{L}_{cross}$.}
	\label{fig:teaserVis}
\end{figure}

%% file: sec/2_related.tex
\section{Related Work}

\noindent\textbf{Referring Expression Comprehension.}
REC aims to ground a language query to the corresponding region onto an image. Most of the advances in REC can be broadly classified into two categories, \ie, two-stage methods and one-stage methods.

Two-stage REC methods~\cite{mao2016generation,wang2016learning,yu2016modeling,liu2019learning,wang2019neighbourhood,yang2019dynamic,yu2018mattnet,hu2017modeling,zhang2018grounding,zhuang2018parallel,liu2019improving,hong2019learning,chen2021ref} adopt the \emph{propose-and-rank} pipeline, where region candidates are firstly generated using unsupervised methods~\cite{plummerCITE2018,wang2019learning} or the pre-trained object detector~\cite{yu2018mattnet,zhang2018grounding}. Then the similarity scores between the sentence and region candidates are computed, which can be discriminatively trained with the binary classification~\cite{wang2019learning,zhang2017discriminative} or maximum-margin ranking~\cite{mao2016generation, nagaraja2016modeling, wang2016learning}.

One-stage REC methods~\cite{yang2019fast,liao2020real,yang2020improving,zhou2021real,luo2020multi,sun2021iterative,huang2021look} get rid of the computation-intensive proposal generation process and aim to directly perform bounding box prediction based on the the language-attended feature maps. The pioneering work \cite{yang2019fast} fuses the language features into the YOLOv3 detector~\cite{redmon2018yolov3} to ground the referred instance. To take one step further, ReSC~\cite{yang2020improving} proposes a recursive sub-query construction framework, which conducts multi-round image-query reasoning and reduces the referring ambiguity. To achieve real-time inference, RCCF~\cite{liao2020real} formulates the REC problem as a correlation filtering process~\cite{bolme2010visual,henriques2014high} by employing the peak value in the correlation heatmap as the center points of the target box. Recently, several works~\cite{deng2021transvg,kamath2021mdetr,zhou2021trar} rely on the Transformer-family structures~\cite{vaswani2017attention} to perform the end-to-end grounding without post-processing procedures.

\noindent\textbf{Video Referring Expression Comprehension.} 
Besides the image REC, there are several works exploring the task in the video domain. Compared to video grounding~\cite{cao2021pursuit,zhang2021cola,cao2021deep,zhang2021synergic,zhang2022unsupervised,cao2022locvtp}, video REC requires both spatial and temporal localization results. According to the specific settings, they can be classified into three threads. 1) Several works~\cite{huang2018finding,yu2017sentence,zhou2018weakly,shi2019not,yang2020weakly} aim to localize all the mentioned objects in the corresponding regions for each frame of the video; 2) There are also works identifying spatial-temporal tubes within untrimmed videos from referring expressions~\cite{zhang2020object,zhang2020does}; 3) {The other works~\cite{sadhu2020video,chen2019weakly,song2021co} localize only the referred object corresponding to the global query sentence semantics within the trimmed videos.}

This paper targets the third setting since it is a more basic and fundamental problem. Chen \emph{et.al}~\cite{chen2019weakly} take the two-stage manner by firstly extracting potential spatio-temporal tubes and then aligning these candidates to the sentence to find the best matching one. \cite{song2021co} proposes a one-stage framework built on YOLO-v3~\cite{redmon2018yolov3}, \ie, it fuses visual-text features and predicts bounding boxes densely at all spatial locations. \cite{sadhu2020video} investigates the role of object relations across both time and space by modeling self-attention with relative position encoding. These methods, however, do not introduce detailed alignment between the mentioned object instances and object areas, making it hard to identify the referent. In this paper, we aim to build this comprehensive alignment for all densely placed instances and achieve more accurate grounding results.

\noindent\textbf{Semantic Correspondence.} 
Establishing correspondences between two images has been investigated extensively for decades~\cite{arandjelovic2016netvlad,NUS_CVPR18_PointNetVLAD,Liu_PAMI11_SiftFlow,UCBerkeley_ECCV20_RANSACFlow}, which is crucial for many vision tasks such as 3D reconstruction~\cite{geiger2011stereoscan}, image retrieval~\cite{chen2021deep}, visual SLAM~\cite{kerl2013dense}, \etc. 

Previous works require explicit supervision and are confined to the single visual modality. In this paper, we aim to learn the correspondence via the latent feature similarity computation in both the inter-modal and cross-modal settings. This simple manner is demonstrated to be effective to capture accurate correspondence under both scenarios.

%% file: sec/3_method.tex
\input{table_figs/figPipeline}


\section{Approach}

This work aims to introduce a dual correspondence network for video REC. We firstly present the backbone of our DCNet including the video-language fusion process in Sec.~\ref{sec:3.1}. Then we elaborate the proposed inter-frame and cross-modality contrastive loss in Sec.~\ref{sec:3.2} and Sec.~\ref{sec:3.3}, respectively. The final training objectives are presented in Sec.~\ref{sec:3.4}.

\subsection{Backbone of DCNet}\label{sec:3.1}

The schematic illustration of our DCNet is illustrated in Figure.~\ref{fig:pipeline}. Given one video-query pair, we firstly feed the video and language modalities to their respective encoders $f_{v}(\cdot)$ and $f_{q}(\cdot)$ to obtain embedded features. Specifically, during each iteration, we randomly sample $T$ video frames and each two sampled frames are with $F$ frame temporal distance. The encoded video feature is represented as $\boldsymbol{v}= \{\boldsymbol{v}^{i}\}_{i=1}^{T}$. The $i^{th}$ frame feature is denoted as $\boldsymbol{v}^{i} = \{\boldsymbol{v}^{i}_{p}\}_{p=1}^{P}$, where $\boldsymbol{v}^{i}_{p} \in \mathbb{R}^{D}$ is the $p^{th}$ patch feature of the $i^{th}$ frame\footnote{The $i^{th}$ frame feature  $\boldsymbol{v}^{i}_{p}$ is a 2-D feature and the $p^{th}$ patch is indexed by the order of \emph{firstly-row-and-then-column}.}. $P$ is the total patch number and $D$ is the feature dimension. The text embedding is represented as $\boldsymbol{q}= \{\boldsymbol{q}^{s}\}_{s=1}^{S}$, where $\boldsymbol{q}^{s} \in \mathbb{R}^{D}$ is the $s^{th}$ word embedding and $S$ is the total word length.

Then the two modality features are fused via a video-language fusion module. Without loss of generality, we utilize the attention mechanism to aggregate the word features for each patch feature following \cite{chen2019weakly,zhang2020does}. Specifically, given $i^{th}$ frame patch feature with the index $p \in [1, P]$, \ie, $\boldsymbol{v}_{p}^{i}$, we compute its attentive scores with each query word $\boldsymbol{q}^{s}, s \in [1, S]$ and apply a $\operatorname{Softmax}$ to normalize\footnote{For conciseness, we omit all the bias terms of linear transformations in this paper.}:
\begin{equation}
e_{p,s}^{i} = \operatorname{Softmax} \left( \mathbf{W}^{\mathrm{T}} \tanh (\mathbf{W}^{v} \boldsymbol{v}_{p}^{i}+\mathbf{W}^{q} \boldsymbol{q}^{s})\right), \label{eq:5}
\end{equation}

\noindent where $\mathbf{W}, \mathbf{W}^{v}, \mathbf{W}^{q} \in \mathbb{R}^{D \times D}$ are learnable parameters. $e_{p,s}^{i}$ is the computed attention weight for patch feature $\boldsymbol{v}_{p}^{i}$ with regard to word feature $\boldsymbol{q}^{s}$. Accordingly, the cross-modal fused features are computed as follows:
\begin{equation}
\boldsymbol{f}^{i}_{p} =\sum_{s=1}^{S} e_{p,s}^{i} \boldsymbol{q}^{s},\label{eq:6}
\end{equation}

\noindent where $\boldsymbol{f}^{i}_{p}$ is the textual-aware video feature for $p^{th}$ patch at $i^{th}$ frame. Thus, the cross-modality fused feature of $i^{th}$ frame is denoted as $\boldsymbol{f}^{i} = \{\boldsymbol{f}^{i}_{p}\}_{p=1}^{P}$. Through this, we characterize the matching behaviors between each patch and the query word features, leading to well-fused cross-modal features.

\subsection{Inter-frame Contrastive Learning}\label{sec:3.2}

We propose to build the patch-wise correspondence and conduct the inter-frame contrastive learning. We contend that introducing such alignment helps incorporate more contextual information and calibrate the frame features to be more temporally consistent. The most obvious challenge is the acquisition of inter-frame correspondence. Instead of directly using the annotated per-frame ground-truth areas as the positive samples, we propose a \emph{dense} alignment manner which computes the similarity scores among all patch pairs.

As shown in Figure.~\ref{fig:pipeline}, we take two encoded image frame features $\boldsymbol{v}^{i}$ and $\boldsymbol{v}^{j}$, $i,j \in [1, T]$ as an example to illustrate the details. For each patch within $\boldsymbol{v}^{i}$, \eg, $\boldsymbol{v}^{i}_{p}, p \in [1, P]$, we firstly computes its cosine similarity with all patch feature in $\boldsymbol{v}^{j}$. The items with top $K_{inter}$ similarity scores are selected and pooled as the positive samples.
\begin{equation}
\boldsymbol{m}^{i}_{p}=\operatorname{avgpool} \Big(\underset{\boldsymbol{v}^{j}_{q}, q\in [1, P]}{\arg \operatorname{topk} } \big(\boldsymbol{v}^{i}_{p} {\cdot} \boldsymbol{v}^{j}_{q}\big)\Big), \label{equ:1}
\end{equation}
\noindent where $\boldsymbol{m}^{i}_{p}$ is the final positive sample for $\boldsymbol{v}^{i}_{p}$ and $\boldsymbol{v}^{j}_{q}$ is the $q^{th}$ patch of $\boldsymbol{v}^{j}$. $(\boldsymbol{u} {\cdot} \boldsymbol{v})=\boldsymbol{u}^{\mathrm{T}} \boldsymbol{v} /\|\boldsymbol{u}\|\|\boldsymbol{v}\|$ represents the cosine similarity between $\ell_{2}$ normalized $\boldsymbol{u}$ and $\boldsymbol{v}$. The selected positive number is defined as $K_{inter} = \max (1, \lfloor {P}/{R_{inter}}\rfloor)$, where $P$ is the total patch number and $1/R_{inter}$ is the sampling ratio.

Based on the established patch-wise correspondence, we perform the inter-frame contrastive learning following the cross-modal InfoNCE~\cite{he2020momentum} loss. 
\begin{equation}
\mathcal{L}_{inter}^{ij}=\frac{1}{P}\sum_{p=1}^{P}-\log \frac{ \exp \Big(\boldsymbol{v}^{i}_{p} {\cdot} \boldsymbol{m}^{i}_{p} / \tau \Big)}{\sum_{q=1}^{P} \exp \left(\boldsymbol{v}^{i}_{p} {\cdot} \boldsymbol{v}^{j}_{q} / \tau\right)}, \label{equ:2}
\end{equation}
\noindent where $\tau$ is the temperature parameter.

\input{table_figs/figInterLossCombine}


To aggregate $\mathcal{L}_{inter}^{ij}$ among all sampled $T$ frames, we propose two potential ways to obtain the final inter-frame contrastive loss $\mathcal{L}_{inter}$. As shown in Figure.~\ref{fig:InterLossCombine}, the \emph{adjacent} strategy only conducts the inter-frame contrastive loss between the nearest frames while the \emph{fully-connected} strategy aggregates frame-wise associations from all the other frames.

\subsection{Cross-modal Contrastive Learning}\label{sec:3.3}

Although REC only requires to ground the object corresponding to the overall semantic meanings, we contend that conducting the fine-grained alignment for each object is crucial to differentiate the referent and the contextual objects, leading to more accurate localization results. Thus we propose to build the alignment for each frame patch and the corresponding words. Similar to the inter-frame contrastive learning in Sec.~\ref{sec:3.2}, the fine-grained \emph{patch-word} alignment is also estimated based on the feature similarity scores. Given the $p^{th}$ patch of the $i^{th}$ frame $\boldsymbol{v}^{i}_{p}$ and the $s$-$th$ word feature $\boldsymbol{q}^{s}$, we compute the similarity scores between them and selected the top $K_{cross}$ responsive items as the positive samples.
\begin{equation}
\boldsymbol{w}^{i}_{p}=\operatorname{avgpool} \Big(\underset{\boldsymbol{q}^{s}, s\in [1, S]}{\arg \operatorname{topk} } \left(\boldsymbol{v}^{i}_{p} {\cdot} \boldsymbol{q}^{s}\right)\Big), \label{equ:3}
\end{equation}
\noindent where $\boldsymbol{w}^{i}_{p}$ is the pooled positive cross-modal features for the image patch $\boldsymbol{v}^{i}_{p}$. $K_{cross} = \max (1, \lfloor {S}/{R_{cross}}\rfloor)$, where $S$ is the total word length and $1/R_{cross}$ is the sampling ratio.

Then we build the  cross-modal contrastive learning based on the mined patch-word correspondence as  follows: 
\begin{equation}
\mathcal{L}_{cross}^{i}=\frac{1}{P}\sum_{p=1}^{P} -\log \frac{ \exp \Big(\boldsymbol{v}^{i}_{p} {\cdot} \boldsymbol{w}^{i}_{p} / \tau \Big)}{\sum_{s=1}^{S} \exp \left(\boldsymbol{v}^{i}_{p} {\cdot} \boldsymbol{q}^{s} / \tau\right)}. \label{equ:4}
\end{equation}

Thus the final cross-modal contrastive loss is represented as $\mathcal{L}_{cross} = \frac{1}{T}\sum_{i=1}^{T} \mathcal{L}_{cross}^{i}$. 

\subsection{Training Objectives}\label{sec:3.4}

Besides the aforementioned inter-frame and cross-modal contrastive loss, we also apply a localization loss $\mathcal{L}_{loc}$ to regress the bounding box towards the ground-truth and a classification loss $\mathcal{L}_{cls}$ to select the best matching prediction term~\cite{song2021co,chen2019weakly}. We apply these two losses based on the cross-modal features illustrated in Sec.~\ref{sec:3.1}. For each spatial position of the cross-modal features, we generate bounding box predictions centered at the current location, together with a confidence score to indicate the probability to be the final grounding output. Specifically, the regression and classification losses are implemented as the MSE (mean square error) loss and the cross-entropy loss, respectively. During inference, the bounding box with the highest confidence is selected as the final prediction. Please refer to \cite{redmon2018yolov3,yang2019fast} for more details.

Integrating the above constraints, final loss function is as follows.
\begin{equation}
\mathcal{L} = \lambda_{loc} \mathcal{L}_{loc} + \lambda_{cls}  \mathcal{L}_{cls} + \lambda_{inter} \mathcal{L}_{inter} + \lambda_{cross} \mathcal{L}_{inter},
\label{eq:9}
\end{equation}
\noindent where $\lambda_{loc}$, $\lambda_{cls}$, $\lambda_{inter}$, and $\lambda_{inter}$ are balance factors.

%% file: table_figs/figPipeline.tex
\begin{figure*}[t]
	\centering
	\includegraphics[width=0.9\textwidth]{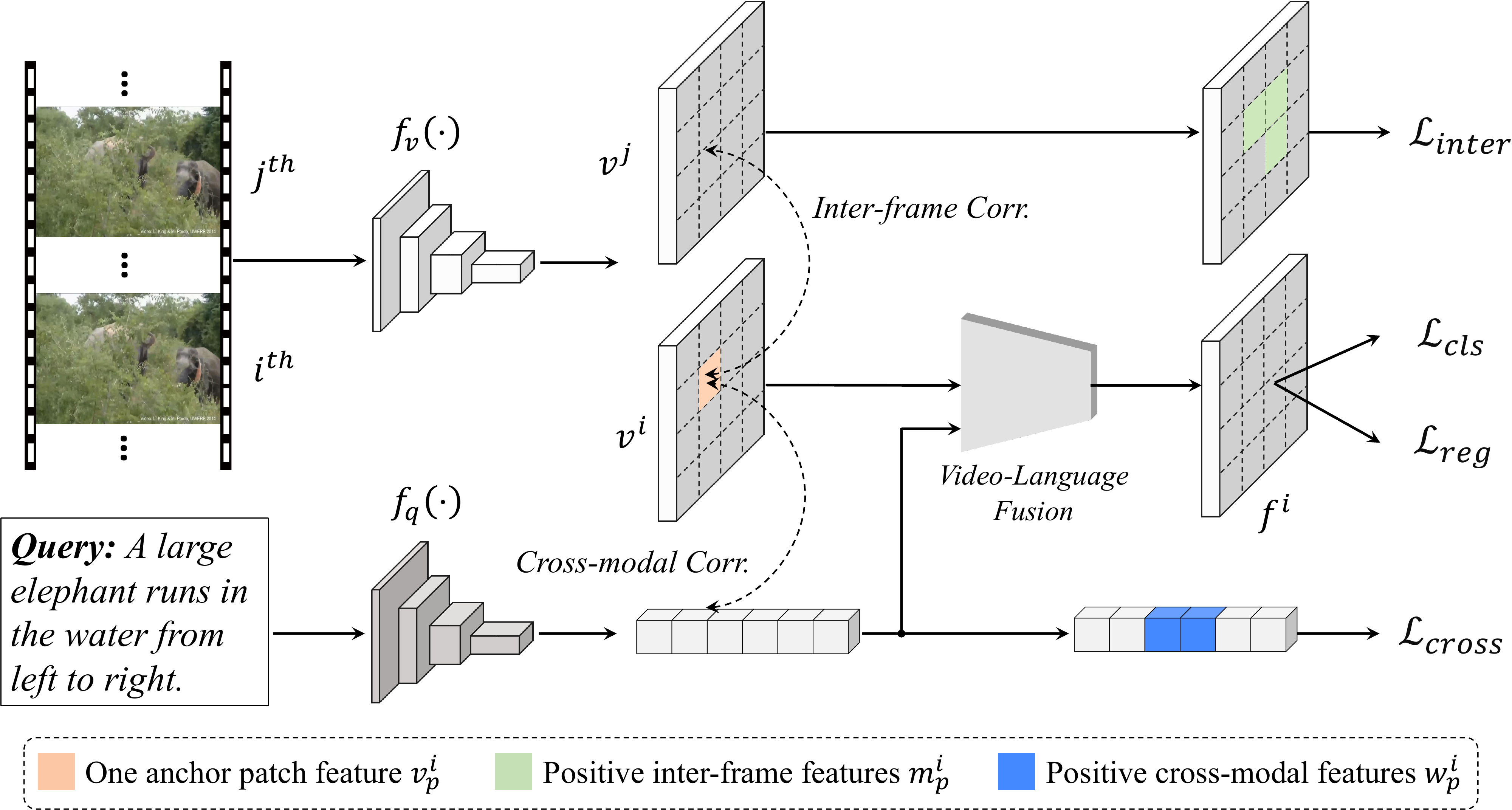}
	\caption{An overview of DCNet. $f_{v}(\cdot)$ and $f_{q}(\cdot)$ are video and language encoders, respectively. Inter-frame correspondence mines the dense patch-wise coherence while cross-modal correspondence finds the detailed patch-word alignment. Note that for clarity, we only present the correspondence discovery for one selected anchor patch $\boldsymbol{v}^{i}_{p}$ (marked by \textcolor[RGB]{252,196,164}{\textbf{orange}} square). Then the discovered inter-frame positive patches $\boldsymbol{m}^{i}_{p}$ are marked by \textcolor[RGB]{200,228,180}{\textbf{green}} while the cross-modal positive words $\boldsymbol{w}^{i}_{p}$ are marked by \textcolor[RGB]{68,140,252}{\textbf{blue}} Based on this, inter-frame and cross-modal contrastive loss $\mathcal{L}_{inter}$, $\mathcal{L}_{cross}$ are applied, \ie, $\boldsymbol{v}^{i}_{p} \leftrightarrow \boldsymbol{m}^{i}_{p}$, $\boldsymbol{v}^{i}_{p} \leftrightarrow \boldsymbol{w}^{i}_{p}$. The video-language fusion module fuses the cross-modal features via the attention mechanism. Based on it, $\mathcal{L}_{cls}$ and $\mathcal{L}_{reg}$ conduct the classification and regression loss for each spatial position of the features. Note that we omit the video-language fusion as well as the classification and regression loss for frame feature $\boldsymbol{v}^{j}$ for clarity.}
	\label{fig:pipeline}
\end{figure*}

%% file: table_figs/figInterLossCombine.tex
\begin{figure}[t]
	\centering
	\includegraphics[width=0.45\textwidth]{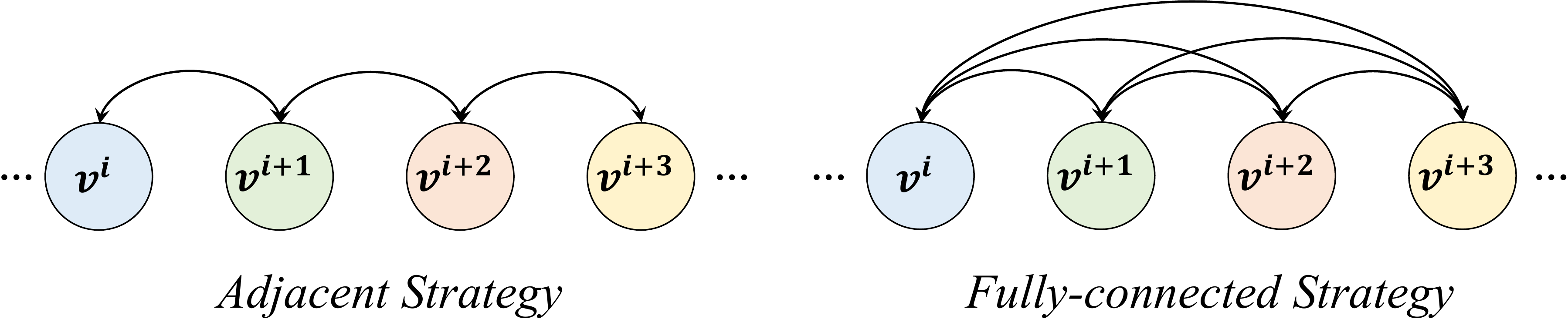}
	\caption{Illustration of two configurations for the inter-frame contrastive learning.}	
	\label{fig:InterLossCombine}
\end{figure}

%% file: sec/4_exp.tex
\section{Experiments}\label{expes}

We first introduce the experimental settings in Sec.~\ref{sec:4_1}. The comparisons with the state-of-the-art methods are discussed in Sec.~\ref{sec:4_2}. Then, we present the model architecture agnostic analysis in Sec.~\ref{sec:4_3}. The extensive exploratory studies on cross-modal correspondence and inter-frame correspondence learning are provided in Sec.~\ref{sec:4_4} and Sec.~\ref{sec:4_5}, respectively. Note that all ablation studies are conducted on the VID-Sentence dataset. Further, we provide the visualization results in Sec.~\ref{sec:4_6}.

\input{exps/4_1_setting}
\input{exps/4_2_SOTA}
\input{exps/4_3_ablation}
\input{exps/4_4_vis}

%% file: exps/4_1_setting.tex
\subsection{Experimental Settings}\label{sec:4_1}
\textbf{Datasets.} We evaluated the performance of REC on two challenging video REC datasets and one image REC dataset. 1) \textbf{VID-Sentence (VID)}~\cite{chen2019weakly}: It is re-labeled by Chen \textit{et al.}~\cite{chen2019weakly} based on the ImageNet video object detection dataset~\cite{russakovsky2015imagenet}. It includes 7,654 trimmed videos with annotated language captions. The size of the vocabulary is 1,823 and the average length of the descriptions is 13.2. Following~\cite{chen2019weakly,song2021co}, we adopt 6,582 instance pairs for training and 536 pairs for testing. 2) \textbf{Lingual OTB99 (LiOTB)}~\cite{li2017tracking}: This dataset is built based on the popular object tracking dataset OTB100~\cite{lu2014online} which contains 100 videos. 
We take the same strategy as in \cite{li2017tracking,song2021co}, splitting the dataset into 51 and 48 instances for training and testing, respectively. 3) \textbf{RefCOCO}~\cite{yu2016modeling}: It consists of 142,210 referring expressions for 50,000 objects in 19,994 images. The average length of each expression is 3.5 words. The dataset is split into train, validation, testA, and testB sets. The testA and testB sets contain images with multiple people and objects, respectively.

\noindent \textbf{Evaluation Metrics.} Following~\cite{yang2020grounding,song2021co,li2022end}, we adopt ``$\text{Accu.}@\alpha$" as the metrics. Specifically, we compute the intersection over union (IoU) for the predicted bounding box and the ground truth per frame. The prediction for one video is considered as positive if the IoU of all frames exceeds the threshold $\alpha$. We set $\alpha$ to 0.4, 0.5, 0.6. Besides, following~\cite{yang2020grounding,song2021co}, we additionally adopt ``success" and ``precision" as the evaluation metrics. The success score is the same as the AUC (area under curve) metric while the precision score represents the ratio of frames where the predicted bounding box falls within a threshold of 20 pixels around the target.

\noindent \textbf{Implementation Details.} We chose Darknet-53~\cite{redmon2018yolov3} pre-trained on MSCOCO~\cite{lin2014microsoft} and BERT~\cite{devlin2018bert} pre-trained on English Wikipedia and Toronto Book Corpus as the visual and language encoder, respectively. The long edge of the input image was resized to the size of 256 and then the whole image was padded to $256 \times 256$. The extracted image feature was with the multi-scale resolution, \ie, $8\times8$,$16\times16$ and $32\times32$. Random crop, affine transformation, horizontal flip, and color jitter were used for image data augmentation. The network was optimized with RMSProp~\cite{tieleman2012lecture} for 10 epochs in video datasets and 100 epochs in image dataset. The initial learning rate was set to 1e-4 and decayed with a polynomial schedule. Experiments were conducted on 4 V100 GPUs and the batch sizes for training VID, LiOTB, and RefCOCO were set to 32, 8, and 32, respectively. We set the balancing weight $\lambda_{loc}$ = 5, $\lambda_{cls}$ = 1, $\lambda_{inter}$ = 1, and $\lambda_{inter} = 1$. During each iteration, The sampled frame number $T$ was set to 4, and the frame distance $F$ was set to 3. The temperature factor $\tau$ used in contrastive learning was set to 0.07 following \cite{wu2018unsupervised,radford2021learning}. The positive sample sampling rate $1/R_{inter}$ and $1/R_{cross}$ are set to $1/8$ and $1/3$, respectively. Note that for the image dataset RefCOCO, we omitted the inter-frame contrastive loss $\mathcal{L}_{inter}$.

%% file: exps/4_2_SOTA.tex
\subsection{Comparisons with State-of-the-Arts} \label{sec:4_2}

\noindent \textbf{Results on Video REC datasets.} The comparison results on VID-Sentence (VID) and Lingual OTB99 (LiOTB) datasets are summarized in Table.~\ref{table:VID} and Table.~\ref{table:LiOTB}, respectively. Specifically, we provide three sets of comparison methods: 1) \emph{Per-frame grounding\footnote{Please refer to \cite{song2021co} and \cite{chen2019weakly} for details.\label{perframe}}}: We apply widely adopted image REC methods including Yang \emph{et.al}~\cite{yang2019fast}, DVSA~\cite{karpathy2015deep}, and GroundeR~\cite{rohrbach2016grounding} for frame-wise grounding. 2) \emph{Per-frame tracking}: We firstly use \cite{yang2019fast} to obtain the tracking template on the first/middle/last/random frame and employ tracker~\cite{li2019siamrpn++} to track the given template. 3) \emph{Other SOTA video REC methods}: WSSTG~\cite{chen2019weakly}, LSAN~\cite{li2017tracking}, and Co-grounding~\cite{song2021co}.

The results on both datasets show that DCNet achieves state-of-the-art performance on all metrics, strongly demonstrating its effectiveness. For example, in Table.~\ref{table:VID}, our DCNet surpasses the previous SOTA method Co-grounding~\cite{song2021co} 2.19\% on Accu.@0.4.

\noindent \textbf{Results on Image REC datasets.} To further investigate the performance in the image domain, we conduct experiments on the widely adopted image REC dataset RefCOCO.

As shown in Table~\ref{table:refcoco}, our DCNet outperforms current SOTA two-stage and one-stage methods. Notably, we even surpass models trained in the multi-task manner using REC and Referring Expression Segmentation (RES) data jointly. For example, we surpass MAttN~\cite{yu2018mattnet} with up to 4.72\% absolute performance improvement. Besides, compared to methods using the visual encoder pre-trained without excluding val/test images (\eg, ReSC~\cite{yang2020improving}, Trar~\cite{zhou2021trar}), our DCNet also achieves better performance.

\input{table_figs/tabVID}


\input{table_figs/tabLiOTB}


\input{table_figs/tabRefCoCo}

%% file: table_figs/tabVID.tex
\begin{table}[t]
\caption{Comparison (\%) with state-of-the-art methods on VID-Sentence dataset.}
\vspace{-1em}
\renewcommand\arraystretch{1.1}
\setlength{\tabcolsep}{5pt}
\label{table:VID}
\begin{center}
\resizebox{\linewidth}{!}{
\begin{tabular}{lx{15}x{15}x{15}x{15}x{15}}
\toprule
\multirow{2}*{\textbf{Method}} & \multicolumn{3}{c}{\textbf{Accu.@}} & \multirow{2}*{\textbf{Succ.}} & \multirow{2}*{\textbf{Prec.}} \\
\cmidrule{2-4}
& 0.4 & 0.5 & 0.6 & &  \\
\midrule
Yang \emph{et al.} (\emph{w/} BERT)~\cite{yang2019fast} & - & 52.39 & - & 42.7 & 37.3 \\
Yang \emph{et al.} (\emph{w/} LSTM)~\cite{yang2019fast} & - & 54.78 & - & 45.1 & 39.3 \\
DVSA~\cite{karpathy2015deep}+Avg & 36.2 & 29.7 & 23.5 & - & - \\
DVSA~\cite{karpathy2015deep}+NetVLAD~\cite{arandjelovic2016netvlad} & 31.2 & 24.8 & 18.5 & - & - \\
DVSA~\cite{karpathy2015deep}+LSTM & 38.2  & 31.2 & 23.5 & - & -  \\
GroundeR~\cite{rohrbach2016grounding}+Avg & 36.7 & 31.9 & 25.0 & - & -  \\
GroundeR~\cite{rohrbach2016grounding}+NetVLAD~\cite{arandjelovic2016netvlad} & 26.1 & 22.2 & 15.1 & - & - \\  
GroundeR~\cite{rohrbach2016grounding}+LSTM & 36.8 & 31.2 & 27.1 & - & - \\
\cdashline{1-6}[2pt/2pt]
First-frame tracking~\cite{li2019siamrpn++} & - & 36.97 & - & 33.4 & 25.0 \\
Middle-frame tracking~\cite{li2019siamrpn++} & - & 44.00 & - & 38.4 & 30.7  \\
Last-frame tracking~\cite{li2019siamrpn++} & - & 36.26 & - & 32.8 & 23.9  \\
Random-frame tracking~\cite{li2019siamrpn++} & - & 40.20 & - & 35.6 & 27.8 \\
\cdashline{1-6}[2pt/2pt]
WSSTG~\cite{chen2019weakly} & 44.60 & 38.20 & 28.90 & - & - \\
Co-grounding~\cite{song2021co} & 63.35  & 60.25 & 53.89 & 49.5 & 46.2  \\
\cdashline{1-6}[2pt/2pt]
\textbf{DCNet (Ours)} & \textbf{65.54} & \textbf{61.69} & \textbf{55.11} & \textbf{51.1} &\textbf{48.3} \\
\bottomrule
\end{tabular}
}
\end{center}
\end{table}

%% file: table_figs/tabLiOTB.tex
\begin{table}[t]
\caption{Comparison (\%) with state-of-the-art methods on LiOTB dataset.}
\vspace{-1em}
\renewcommand\arraystretch{1.1}
\label{table:LiOTB}
\begin{center}
\begin{tabular}{lccccc}
\toprule
\textbf{Method} & \textbf{Accu.@0.5} & \textbf{Succ.} & \textbf{Prec.} \\
\midrule
Yang \emph{et al.} (\emph{w/} BERT)~\cite{yang2019fast} & 49.13 & 35.8 & 46.8  \\
Yang \emph{et al.} (\emph{w/} LSTM)~\cite{yang2019fast} & 49.16 & 33.3 & 41.4 \\
\cdashline{1-4}[2pt/2pt]
First-Frame Tracking~\cite{li2019siamrpn++} & 50.93 & 39.1 & 48.2\\
Middle-Frame Tracking~\cite{li2019siamrpn++} & 43.08 & 35.6 & 42.1\\
Last-Frame Tracking~\cite{li2019siamrpn++} & 44.17 & 32.7 & 39.1\\
Random-Frame Tracking~\cite{li2019siamrpn++} & 25.16 & 28.8 & 32.9\\
\cdashline{1-4}[2pt/2pt]
LSAN~\cite{li2017tracking}  & - & 25.9 & - & \\
Co-grounding~\cite{song2021co} & 52.26 & 39.2 & 50.0\\
\cdashline{1-4}[2pt/2pt]
DCNet (Ours) & \textbf{53.21} & \textbf{41.0}& \textbf{51.7} \\
\bottomrule
\end{tabular}
\end{center}
\end{table}

%% file: table_figs/tabRefCoCo.tex
\begin{table}[t]
\caption{Comparison (\%) with state-of-the-art methods on RefCOCO dataset evaluated with Accu.@0.5. RN101 and DN53 denotes ResNet101~\cite{he2016deep} and DarkNet53~\cite{redmon2018yolov3}, respectively. $\dagger$ denote the multi-task training using REC and Referring Expression Segmentation (RES) data jointly. Visual encoders of models with $^\ddag$ are trained without excluding val/test images of RefCOCO.}
\vspace{-1em}
\renewcommand\arraystretch{1.1}
\setlength{\tabcolsep}{5pt}
\label{table:refcoco}
\begin{center}
\resizebox{0.9\linewidth}{!}{
\begin{tabular}{lcccc}
\toprule
\textbf{Method} & \textbf{Vis.Enc.} & \textbf{val} & \textbf{testA} & \textbf{testB} \\
\midrule
\multicolumn{5}{l}{\emph{\textbf{Two Stage:}}} \\
\midrule
VC~\cite{zhang2018grounding} & VGG16 & - & 73.33 & 67.44 \\
ParalAttn~\cite{zhuang2018parallel} & VGG16 & - & 75.31 & 65.52 \\
MAttN$^\dagger$~\cite{yu2018mattnet} & RN101 & 76.40 & 80.43 & 69.28 \\
CM-Att-Erase~\cite{liu2019improving} & RN101 & 78.35 & 83.14 & 71.32 \\
DGA~\cite{yang2019dynamic} & VGG16 & - & 78.42 & 65.53 \\
RvG-Tree~\cite{hong2019learning} & RN101 & 75.06 & 78.61 & 69.85 \\
NMTree$^\dagger$~\cite{liu2019learning} & RN101 & 76.41 & 81.21 & 70.09 \\
Ref-NMS (CM-A-E)~\cite{chen2021ref} & RN101 & 80.70 & 84.00 & 76.04 \\
\midrule
\multicolumn{5}{l}{\emph{\textbf{One Stage:}}} \\
\midrule
RealGIN~\cite{zhou2021real} & DN53 & 77.25 & 78.70 & 72.10 \\
FAOA~\cite{yang2019fast} &  DN53 & 71.15 & 74.88 & 66.32 \\
RCCF~\cite{liao2020real} & DLA34 & - & 81.06 & 71.85 \\
MCN~\cite{luo2020multi} & DN53  &  80.08 & 82.29 & 74.98 \\
ReSC$^\ddag$~\cite{yang2020improving} & DN53 & 77.63 & 80.45 & 72.30 \\
LBYL~\cite{huang2021look} & DN53 & 79.67 & 82.91 & 74.15 \\
TransVG~\cite{deng2021transvg} & RN101 & 81.02 & 82.72 & 78.35 \\
Trar$^\ddag$~\cite{zhou2021trar} & DN53 & - & 81.40 & \textbf{78.60}  \\
Co-grounding~\cite{song2021co} & DN53 & 77.65 & 80.75 & 73.37 \\
DCNet (Ours) & DN53 & \textbf{81.12} & \textbf{83.12} & 77.53 \\
\bottomrule
\end{tabular}
}
\end{center}
\end{table}

%% file: exps/4_3_ablation.tex
\subsection{Architecture Agnostic}\label{sec:4_3}

\input{table_figs/tabMergeAbla}

\input{table_figs/tabLoss}

\noindent \textbf{Loss Component Ablations.} We ablate the loss components of our DCNet to see the difference. As shown in Table.~\ref{table:lossAbla}, both the inter-frame and cross-modal contrastive losses $\mathcal{L}_{inter}$, $\mathcal{L}_{cross}$ are crucial. For example, when removing $\mathcal{L}_{inter}$ (mode \#2 in Table.~\ref{table:lossAbla}), Accu.@0.4 drops by 4.55\% compared to the full version (mode \#1). 

\input{table_figs/tabPlug}

\noindent \textbf{Extending to existing methods.} Our inter-frame and cross-modal contrastive losses $\mathcal{L}_{inter}$ and $\mathcal{L}_{cross}$ are annotation-free and can be seamlessly inserted into existing video REC methods. As shown in Table.~\ref{table:plug}, we select two representative methods including the one-stage method \cite{song2021co} and the two-stage method \cite{chen2019weakly} as baselines.

The results in Table.~\ref{table:plug} show that both $\mathcal{L}_{inter}$ and $\mathcal{L}_{cross}$ benefit the two baseline models. For example, on top of \cite{song2021co}, $\mathcal{L}_{inter}$ and $\mathcal{L}_{cross}$ lead to 0.78\% and 1.57\% performance boost on Accu.@0.4, respectively. These results demonstrate that our inter-frame and cross-modal contrastive losses are general and compatible with various video REC methods.

\subsection{Ablations on Inter-frame Correspondence\label{ablations}}\label{sec:4_4}

\noindent \textbf{Sparse \emph{v.s.} Dense Correspondence.} We design the inter-frame contrastive learning to build the \emph{dense} patch-wise correspondence between frames. As illustrated in Figure.~\ref{fig:sparseDense}, there exists a simple manner to conduct inter-frame alignment between annotated ground-truth, which is called \emph{sparse} alignment here. We conduct comparison experiments for both the dense and sparse alignment manners. 

As shown in Table.~\ref{table:dense}, our proposed dense alignment outperforms the sparse one in all three metrics. Besides, our dense alignment requires no supervised annotations while the sparse alignment needs ground-truth annotations.

\noindent \textbf{Multi-frame Aggregation Manner.} As shown in Figure~\ref{fig:InterLossCombine}, we design two potential ways to aggregate the total $T$ frames in the inter-frame contrastive learning. Specifically, the \emph{adjacent} strategy only considers the nearest frames while the \emph{fully-connected} strategy aggregates frame-wise associations from all the other frames.

The results in Table~\ref{table:aggerate} show that the fully-connected aggregation manner slightly outperforms the adjacent aggregation one. For example on Accu.@0.4, the former achieves 0.08\% boost compared to the latter one. Considering the computational costs of the adjacent aggregation are dramatically lower than that of the fully-connected one ($\mathcal{O}(T)$ \emph{v.s.} $\mathcal{O}(T^2)$), we adopt the adjacent aggregation manner.

\noindent \textbf{Ablations of the Sample Distance $F$.} We conduct the ablation studies on the sampled frame distance $F$ to determine the optimum choice. From the results in Table.~\ref{table:distance}, we can observe that the optimum performance is achieved when setting $F = 3$. Too large or too small $F$ value will both lead to the performance degradation. 

\noindent \textbf{Ablations of the Positive Sample Ratio $1/R_{inter}$.} We conduct ablative experiments on the positive sample ratio $1/R_{inter}$ in the inter-frame contrastive learning. The results in Table.~\ref{table:topk1} show that the performance saturates at $1/R_{inter}=1/8$. This may be because too few patch-wise pairs are unable to establish accurate correspondences while setting too many positive samples makes it hard to find the most informative pairs.

\subsection{Ablations on Cross-modal Correspondence\label{ablations}}\label{sec:4_5}

\input{table_figs/figVisStable}

\input{table_figs/figWordSim}


Cross-modal contrastive loss aims to capture the fine-grained patch-word alignment. In this section, we offer more potential correspondence discovery solutions to find the most suitable one.

\noindent \textbf{Cross-modal Correspondence Discovery.} In Sec.~\ref{sec:3.3}, we propose to find the cross-modal correspondence by selecting the most similar $K_cross$ words for each patch region. Here we refer to it \emph{patch-topk}. We provide two comparison discovery strategies: 1) \emph{word-topk}: we select top-$K_{cross}$ most responsive patches for each word; 2) \emph{random}: we randomly select $K_{cross}$ for each patch feature.

As shown in Table~\ref{table:corresponce}, the \emph{random} selection strategy performs the worst with Accu.@0.4 only reaching 61.35\%. Besides, our \emph{patch-topk} discovery strategy outperforms the \emph{word-topk} manner. For example on Accu.@0.4, the performance of \emph{word-topk} is 1.41\% absolute lower than \emph{patch-topk}. We conjecture that this may be because the  \emph{word-topk} matching manner enforces each word to be mapped close to some patch features. However, there exist words without concrete meanings (\eg, articles or pronouns) and introducing them into matched pairs is meaningless.

\noindent \textbf{Ablations of the Positive Sample Ratio $1/R_{cross}$.} We ablate on $1/R_{cross}$ to determine the optimum hyper-parameter. As shown in Table.~\ref{table:topk2}, our DCNet achieves the best performance when setting $1/R_{cross}=1/3$. Too many or too few positive sample pairs both degrade the performance.

%% file: table_figs/tabMergeAbla.tex
\begin{table*}[h]
\centering
\caption{(a) Comparisons of different alignment manners in inter-frame contrastive learning. \emph{Dense}: our proposed patch-wise correspondence; \emph{Sparse}: alignment conducted between annotated bounding box features. Refer to Figure.~\ref{fig:sparseDense} for schematic illustrations. (b) Comparisons of different aggregation manners. \emph{adjacent}: aggregating from the nearest frames; \emph{fc (fully-connected)}: aggregating from all the other frames. Refer to Figure.~\ref{fig:InterLossCombine} for schematic illustrations. (c) Ablations of the sampled frame distance $F$. (d) Ablations of the positive sample ratio $1/R_{inter}$. (e) Ablations of cross-modal correspondence discovery strategies. (f) Ablations of the positive sample ratio $1/R_{cross}$.}
\vspace{-0.5em}
\renewcommand\arraystretch{1.1}
\setlength{\tabcolsep}{5pt}
    \begin{subtable}[h]{0.32\textwidth}
        \centering
        \begin{tabular}{cccc}
        \toprule
        \textbf{Mode}  & \textbf{Accu.@0.4} & \textbf{@0.5} & \textbf{@0.6} \\
        \midrule
        \emph{Dense} & \textbf{65.54} & \textbf{61.69} & \textbf{55.11}\\
        \emph{Sparse} & 64.15 & 61.07 & 54.12\\
        \bottomrule
        \multicolumn{4}{c}{~}\\ 
        \end{tabular}
        \caption{}
        \label{table:dense}
     \end{subtable}\hfill
    \begin{subtable}[h]{0.32\textwidth}
        \centering
        \begin{tabular}{cccc}
        \toprule
        \textbf{Mode}  & \textbf{Accu.@0.4} & \textbf{@0.5} & \textbf{@0.6} \\
        \midrule
        \emph{adjacent} & 65.54 & 61.69 & 55.11 \\
        \emph{fc} & 65.62 & 61.83 & 55.26 \\
        \bottomrule
        \multicolumn{4}{c}{~}\\ 
        \end{tabular}
        \caption{}
        \label{table:aggerate}
     \end{subtable}\hfill
    \begin{subtable}[h]{0.32\textwidth}
        \centering
        \begin{tabular}{cccc}
        \toprule
        \textbf{$F$}  & \textbf{Accu.@0.4} & \textbf{@0.5} & \textbf{@0.6} \\
        \midrule
        1    & 65.18 & 61.37 & 54.14 \\
        3    & \textbf{65.54} & \textbf{61.69} & \textbf{55.11} \\
        5    & 64.03 & 60.44 & 53.46 \\
        \bottomrule
        \end{tabular}
        \caption{}
        \label{table:distance}
     \end{subtable}
     \begin{subtable}[h]{0.32\textwidth}
        \centering
        \begin{tabular}{cccc}
        \toprule
        \textbf{$1/R_{inter}$}  & \textbf{Accu.@0.4} & \textbf{@0.5} & \textbf{@0.6} \\
        \midrule
        1/12 & 64.81 & 61.02 & 54.68 \\
        1/8 & \textbf{65.54} & \textbf{61.69} & \textbf{55.11} \\
        1/4 & 61.99 & 60.64 & 53.89 \\
        \bottomrule
        \end{tabular}
        \caption{}
        \label{table:topk1}
     \end{subtable}\hfill
    \begin{subtable}[h]{0.35\textwidth}
        \centering
        \begin{tabular}{cccc}
        \toprule
        \textbf{Mode}  & \textbf{Accu.@0.4} & \textbf{@0.5} & \textbf{@0.6} \\
        \midrule
        \emph{random}     & 61.35 & 59.64 & 51.26 \\
        \emph{word-topk}  & 64.13 & 61.27 & 54.66 \\
        \emph{patch-topk} & \textbf{65.54} & \textbf{61.69} & \textbf{55.11} \\
        \bottomrule
        \end{tabular}
        \caption{}
        \label{table:corresponce}
    \end{subtable}
    \begin{subtable}[h]{0.32\textwidth}
        \centering
        \begin{tabular}{cccc}
        \toprule
        \textbf{$1/R_{cross}$}  & \textbf{Accu.@0.4} & \textbf{@0.5} & \textbf{@0.6} \\
        \midrule
        1/5 & 62.72 & 59.04 & 52.11 \\
        1/3 & \textbf{65.54} & \textbf{61.69} & \textbf{55.11} \\
        1/2 & 63.28 & 60.36 & 53.94 \\
        \bottomrule
        \end{tabular}
        \caption{}
        \label{table:topk2}
     \end{subtable}
\end{table*}

%% file: table_figs/tabLoss.tex
\begin{table}[t]
\vspace{-1em}
\caption{Ablations of loss components.}
\vspace{-1em}
\renewcommand\arraystretch{1.1}
\label{table:lossAbla}
\begin{center}
\begin{tabular}{cccccc}
\toprule
\textbf{Mode} & \textbf{$\mathcal{L}_{inter}$} & \textbf{$\mathcal{L}_{cross}$} & \textbf{Accu.@0.4} & \textbf{@0.5} & \textbf{@0.6} \\
\midrule
\#1  & \ding{51} & \ding{51} & \textbf{65.54}  & \textbf{61.69} & \textbf{55.11} \\
\#2 &  \ding{51} & \ding{55} & 60.99  & 59.64 & 52.89 \\
\#3 &  \ding{55} & \ding{51} & 63.13  & 60.09 & 53.05 \\
\#4 &  \ding{55} & \ding{55} & 57.86  &54.78  & 49.35 \\
\bottomrule
\end{tabular}
\end{center}
\end{table}

%% file: table_figs/tabPlug.tex
\begin{table}[t]
\vspace{-0.5em}
\caption{Architecture Agnostic analysis. We insert the proposed inter-frame and cross-modal contrastive losses $\mathcal{L}_{cross}$, $\mathcal{L}_{inter}$ to existing video REC methods~\cite{song2021co,chen2019weakly}.}
\vspace{-1em}
\renewcommand\arraystretch{1.1}
\label{table:plug}
\begin{center}
\begin{tabular}{lccc}
\toprule
\textbf{Method}  & \textbf{Accu.@0.4} & \textbf{@0.5} & \textbf{@0.6} \\
\midrule
Co-grounding~\cite{song2021co}        & 63.35 & 60.25 & 53.89 \\
\quad+$\mathcal{L}_{inter}$   & 64.13 & 61.24 & 54.09\\
\quad+$\mathcal{L}_{cross}$   & 64.92 & 61.24 & 54.21\\
\quad+$\mathcal{L}_{cross}$+ $\mathcal{L}_{inter}$ & 65.58 & 61.73 & 55.41\\
\midrule
WSSTG~\cite{chen2019weakly}         & 44.60 & 38.20 & 28.90 \\
\quad+$\mathcal{L}_{inter}$   & 46.81 & 39.92 & 30.64 \\
\quad+$\mathcal{L}_{cross}$   & 47.14 & 40.41 & 31.57 \\
\quad+$\mathcal{L}_{cross}$+ $\mathcal{L}_{inter}$  & 48.66 & 42.38 & 32.83 \\
\bottomrule
\end{tabular}
\end{center}
\vspace{-1em}
\end{table}

%% file: table_figs/figVisStable.tex
\begin{figure*}[t]
	\centering
	\includegraphics[width=0.9\textwidth]{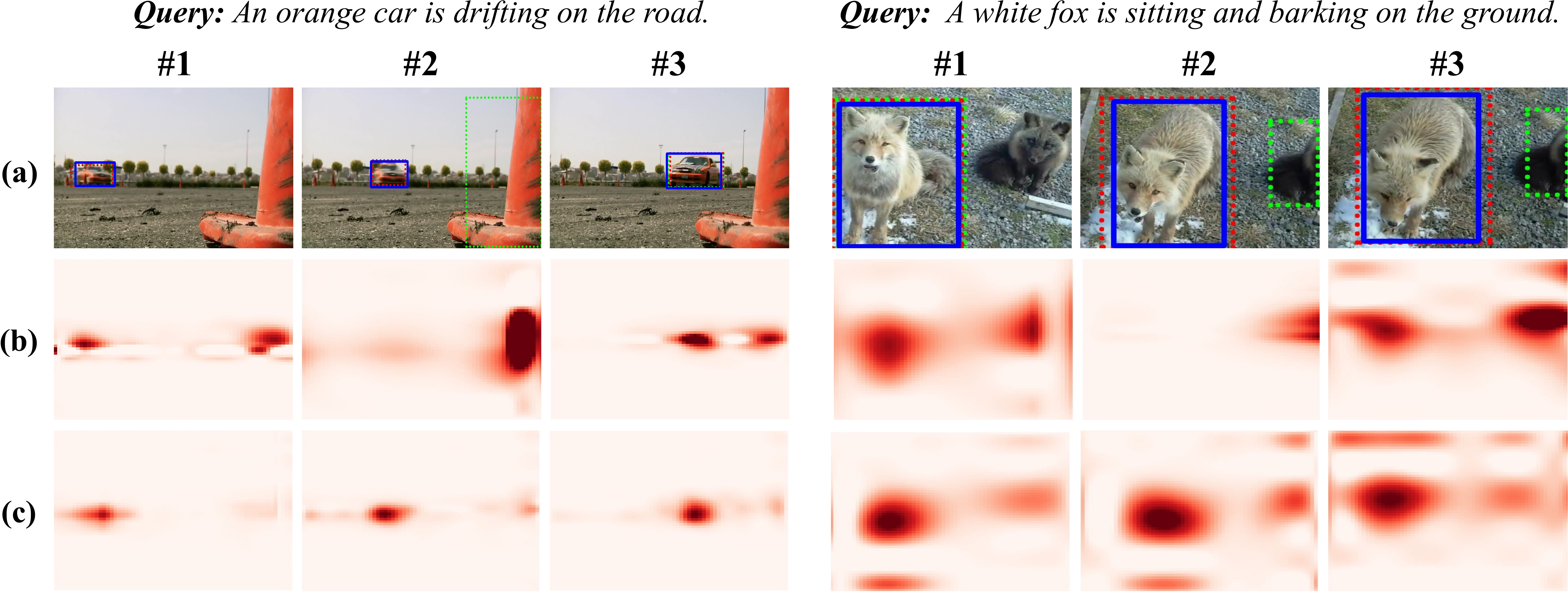}
	\caption{(a) Input image frames. Localization results of DCNet \emph{w/o} and \emph{w/} the inter-frame contrastive loss $\mathcal{L}_{inter}$ are marked in \textcolor{green}{green} and \textcolor{red}{red}, respectively. Ground-truth annotations are marked in \textcolor{blue}{blue}. (b) Confidence score maps  \emph{w/o}  $\mathcal{L}_{inter}$. (c) Confidence score maps \emph{w/} $\mathcal{L}_{inter}$. The score maps of DCNet \emph{w/} $\mathcal{L}_{inter}$ are more stable across video frames than that of DCNet \emph{w/o} $\mathcal{L}_{inter}$. Pay special attention to the \#2 frame of the two examples.}
	\label{fig:VisStable}
\end{figure*}

%% file: table_figs/figWordSim.tex
\begin{figure}[t]
	\centering
	\includegraphics[width=0.4\textwidth]{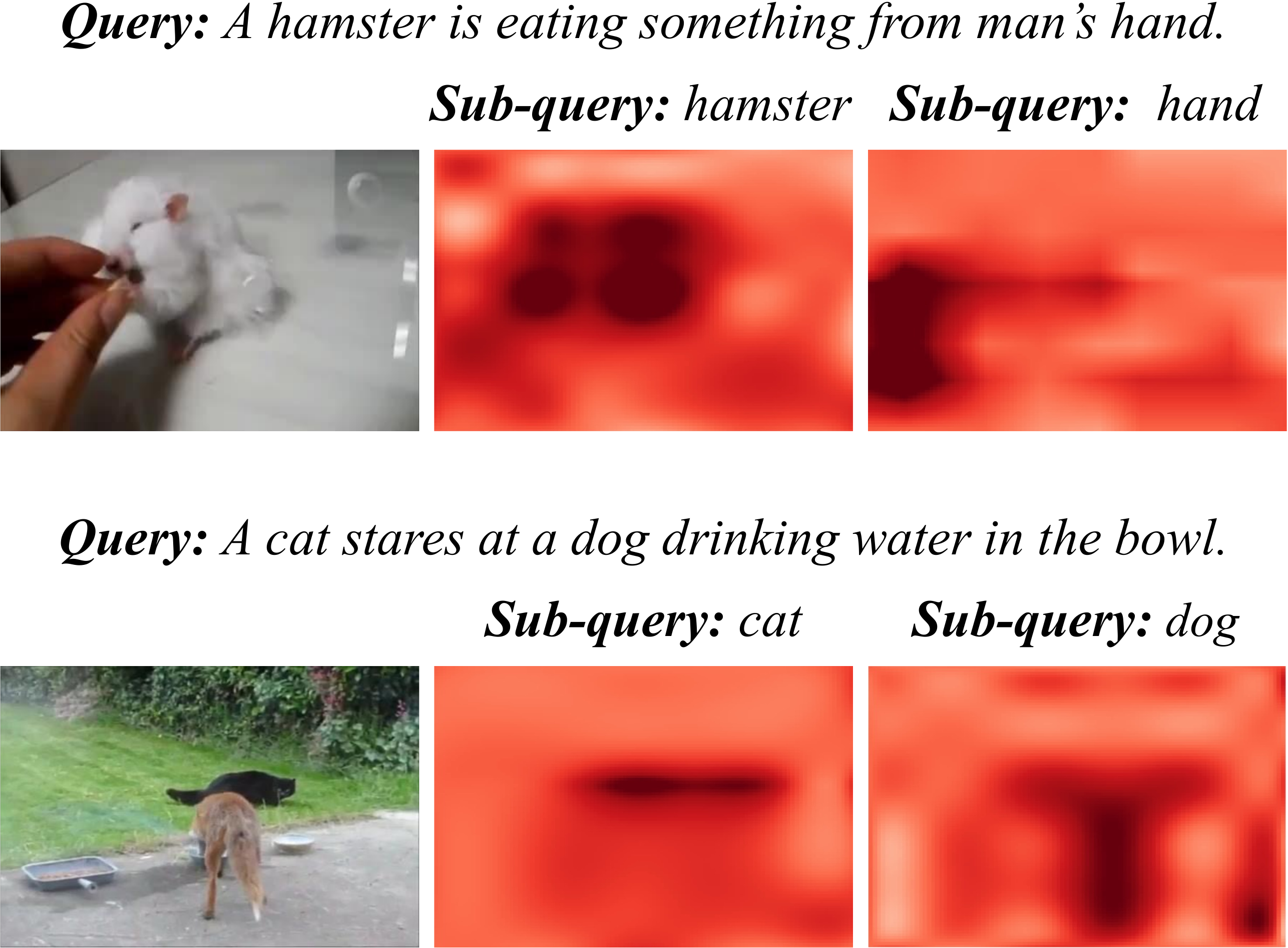}
	\vspace{-1em}
	\caption{Visualizations of the cross-modal alignment. \emph{Sub-query} is the specific word taken from the full \emph{query}. The response map is generated by computing the cosine similarity between the sub-query and the input image.}
	\label{fig:wordSim}
\end{figure}

%% file: exps/4_4_vis.tex
\subsection{Visualizations} \label{sec:4_6}

\noindent \textbf{Visualizations of Confidence Score Maps with (\emph{w/}) and without (\emph{w/o}) $\mathcal{L}_{inter}$.} We visualize the classification confidence score maps for DCNet \emph{w/} and \emph{w/o} $\mathcal{L}_{inter}$. We present two video examples and we select three consecutive frames for each video.

As shown in Figure.~\ref{fig:VisStable}, DCNet \emph{w/o} $\mathcal{L}_{inter}$ leads to unstable and inconsistent feature responses. For example in Figure.~\ref{fig:VisStable} (left), due to the motion blur in frame \#2, DCNet \emph{w/o} $\mathcal{L}_{inter}$ falsely localizes the interfering object in front (marked in \textcolor{green}{green}). In contrast, introducing $\mathcal{L}_{inter}$ helps rectify the localization results in frame \#2, leading to more stable results across frames.

\noindent \textbf{Visualizations of Cross-Modal Alignment.} The cross-modal contrastive loss $\mathcal{L}_{cross}$ builds fine-grained patch-word alignment. To prove this, we select specific words in the language query (denoted as \emph{sub-query} in Figure.~\ref{fig:wordSim})  and compute its cosine similarity with all patches within the frame.

As shown in Figure.~\ref{fig:wordSim}, the similarity distribution focuses more on the areas semantically corresponding to the sub-query. These results demonstrate that our cross-modal contrastive loss $\mathcal{L}_{cross}$ can effectively model the patch-word alignment.

%% file: sec/5_con.tex
\section{Conclusions}
In this paper, we summarized two common problems in video REC, \ie, inconsistent localization and confusion between the referent and contextual objects. To this end, we proposed DCNet which introduces the inter-frame patch-wise and cross-modal patch-word correspondence without the reliance on annotations. Specifically, we estimated the patch-wise and patch-word alignment by computing the cosine similarity in feature space. Then the most responsive pairs were selected as positive samples, where the corresponding inter-frame and cross-modal contrastive learning were applied. Extensive experimental results on both video and image REC datasets have demonstrated the effectiveness of our proposed DCNet. 

\noindent \textbf{Acknowledgements.} This paper was partially supported by NSFC (No: 62176008) and Shenzhen Science \& Technology Research Program (No:GXWD20201231165807007-20200814115301001).

%% file: sec/6_appendix.tex
\section{Appendix}
This supplementary document provides more visualization results. Firstly, we provide qualitative results of the confidence maps with (\emph{w/}) and without (\emph{w/o}) the inter-frame contrastive loss $\mathcal{L}_{inter}$. Then, we show the cross-modal alignment established by our cross-modal contrastive learning.

\noindent \textbf{Visualizations of Confidence Score Maps with (\emph{w/}) and without (\emph{w/o}) $\mathcal{L}_{inter}$.} We visualize the classification confidence score maps for DCNet \emph{w/} and \emph{w/o} $\mathcal{L}_{inter}$. We present three video examples and we select three consecutive frames for each video.

As shown in Figure.~\ref{fig:1}, Figure.~\ref{fig:2} and Figure.~\ref{fig:3}, DCNet \emph{w/o} $\mathcal{L}_{inter}$ leads to unstable and inconsistent feature responses. In contrast, introducing $\mathcal{L}_{inter}$ helps regularize the feature response and rectify the localization results, leading to more stable results across frames.

\noindent \textbf{Visualizations of Cross-Modal Alignment.} The cross-modal contrastive loss $\mathcal{L}_{cross}$ builds fine-grained patch-word alignment. To prove this, we select specific words in the language query (denoted as \emph{sub-query} in Figure.~\ref{fig:4} and Figure.~\ref{fig:5}) and compute its cosine similarity with all patches within the frame.

As shown in Figure.~\ref{fig:4} and Figure.~\ref{fig:5}, the similarity distribution focuses more on the areas semantically corresponding to the sub-query. These results demonstrate that our cross-modal contrastive loss $\mathcal{L}_{cross}$ can effectively model the patch-word alignment.

\input{Appendix/fig1}

\input{Appendix/fig2}
\input{Appendix/fig3}
\input{Appendix/fig4}
\input{Appendix/fig5}

%% file: Appendix/fig1.tex
\begin{figure*}[t]
	\centering
	\begin{subfigure}[b]{0.45\textwidth}
		\centering
		\includegraphics[width=\textwidth]{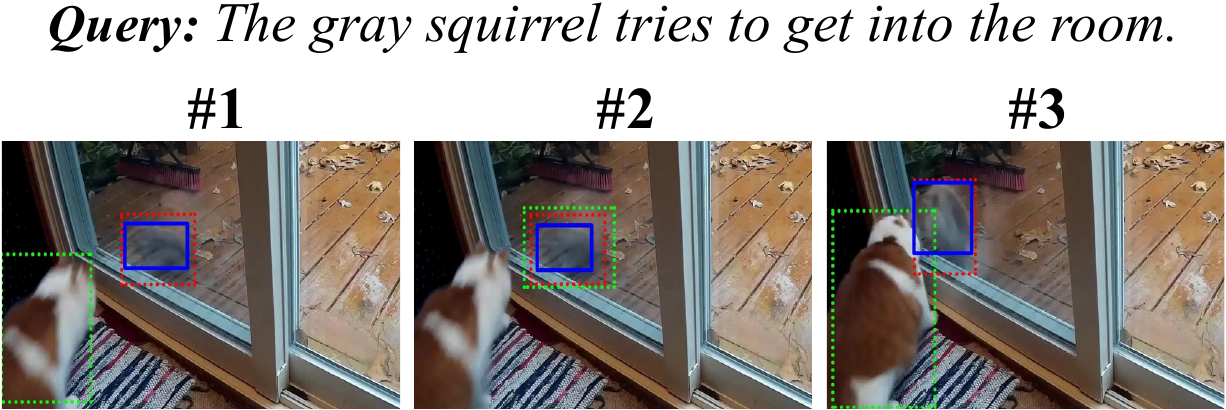}
		\caption{}
		\label{fig:1a}
	\end{subfigure}
	
	\begin{subfigure}[b]{0.45\textwidth}
		\centering
		\includegraphics[width=\textwidth]{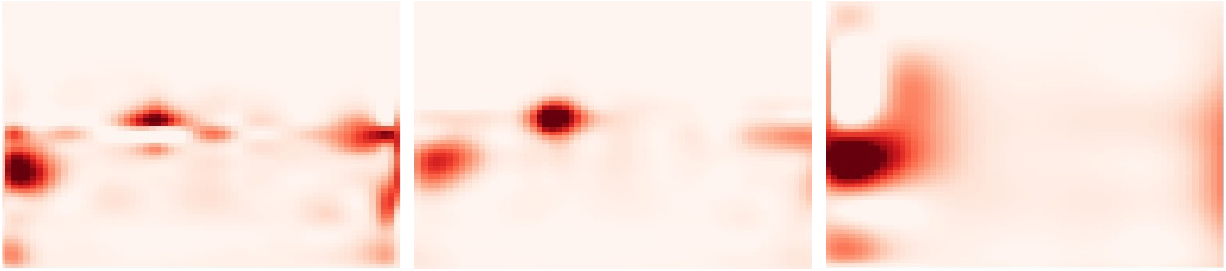}
		\caption{}
		\label{fig:1b}
	\end{subfigure}
	
	\begin{subfigure}[b]{0.45\textwidth}
		\centering
		\includegraphics[width=\textwidth]{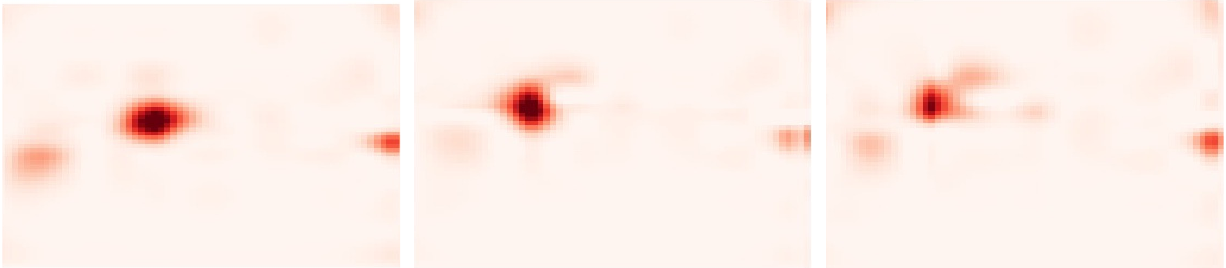}
		\caption{}
		\label{fig:1c}
	\end{subfigure}
	\caption{(a) Input image frames. Localization results for our DCNet \emph{w/o} and \emph{w/} the inter-frame contrastive loss $\mathcal{L}_{inter}$ are marked in \textcolor{green}{green} and \textcolor{red}{red}, respectively. Ground-truth annotations are marked in \textcolor{blue}{blue}. (b) Confidence score maps \emph{w/o}  $\mathcal{L}_{inter}$. (c) Confidence score maps \emph{w/} $\mathcal{L}_{inter}$. The score maps with $\mathcal{L}_{inter}$ are more stable across video frames while the score maps without $\mathcal{L}_{inter}$ are inconsistent. Pay special attention to the \#1 and \#3 frames.}
	\label{fig:1}
\end{figure*}

%% file: Appendix/fig2.tex
\begin{figure*}[t]
	\centering
	\begin{subfigure}[b]{0.45\textwidth}
		\centering
		\includegraphics[width=\textwidth]{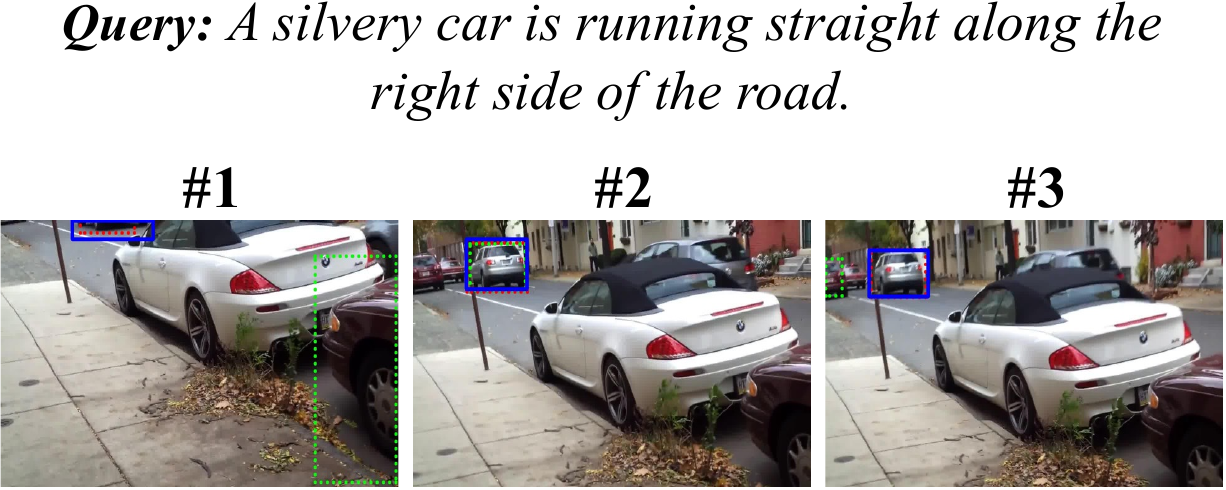}
		\caption{}
		\label{fig:2a}
	\end{subfigure}
	
	\begin{subfigure}[b]{0.45\textwidth}
		\centering
		\includegraphics[width=\textwidth]{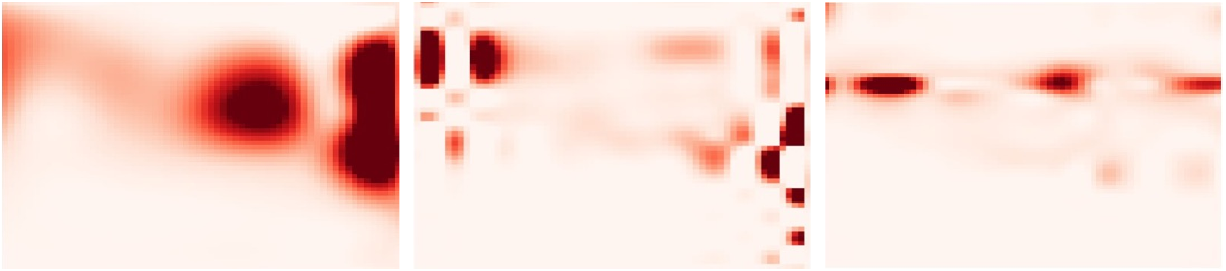}
		\caption{}
		\label{fig:2b}
	\end{subfigure}
	
	\begin{subfigure}[b]{0.45\textwidth}
		\centering
		\includegraphics[width=\textwidth]{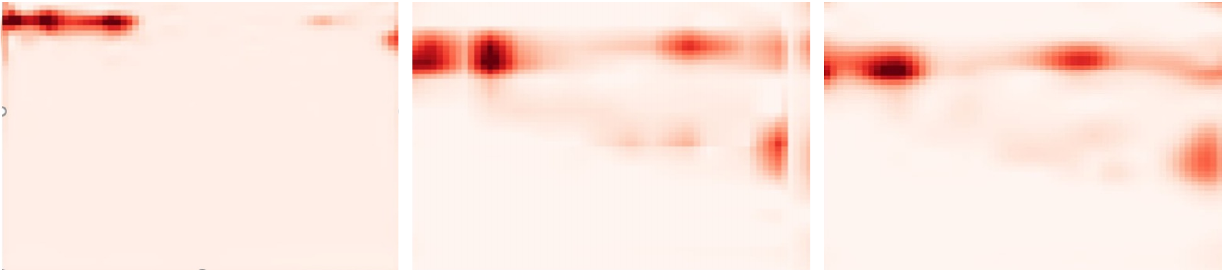}
		\caption{}
		\label{fig:2c}
	\end{subfigure}
	\caption{(a) Input image frames. Localization results for our DCNet \emph{w/o} and \emph{w/} the inter-frame contrastive loss $\mathcal{L}_{inter}$ are marked in \textcolor{green}{green} and \textcolor{red}{red}, respectively. Ground-truth annotations are marked in \textcolor{blue}{blue}. (b) Confidence score maps  \emph{w/o}  $\mathcal{L}_{inter}$. (c) Confidence score maps \emph{w/} $\mathcal{L}_{inter}$. The score maps with $\mathcal{L}_{inter}$ are more stable across video frames while the score maps without $\mathcal{L}_{inter}$ are inconsistent. Pay special attention to the \#1 and \#3 frames.}
	\label{fig:2}
\end{figure*}

%% file: Appendix/fig3.tex
\begin{figure*}[t]
	\centering
	\begin{subfigure}[b]{0.45\textwidth}
		\centering
		\includegraphics[width=\textwidth]{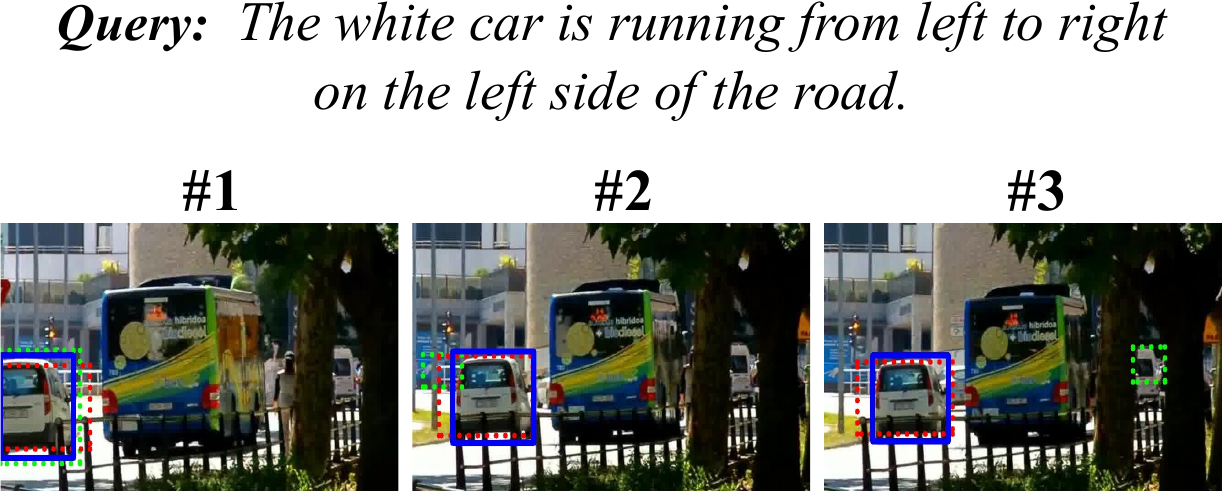}
		\caption{}
		\label{fig:3a}
	\end{subfigure}
	
	\begin{subfigure}[b]{0.45\textwidth}
		\centering
		\includegraphics[width=\textwidth]{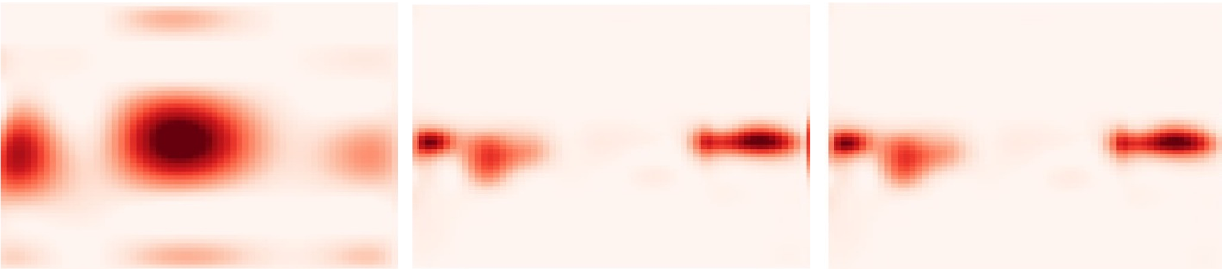}
		\caption{}
		\label{fig:3b}
	\end{subfigure}
	
	\begin{subfigure}[b]{0.45\textwidth}
		\centering
		\includegraphics[width=\textwidth]{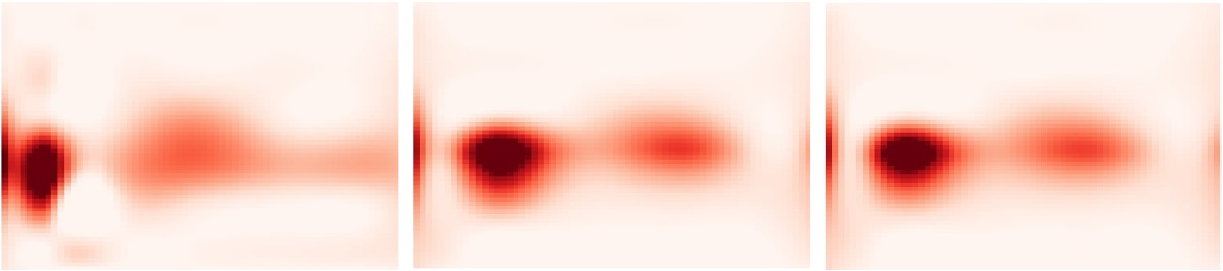}
		\caption{}
		\label{fig:3c}
	\end{subfigure}
	\caption{(a) Input image frames. Localization results for our DCNet \emph{w/o} and \emph{w/} the inter-frame contrastive loss $\mathcal{L}_{inter}$ are marked in \textcolor{green}{green} and \textcolor{red}{red}, respectively. Ground-truth annotations are marked in \textcolor{blue}{blue}. (b) Confidence score maps  \emph{w/o}  $\mathcal{L}_{inter}$. (c) Confidence score maps \emph{w/} $\mathcal{L}_{inter}$. The score maps with $\mathcal{L}_{inter}$ are more stable across video frames while the score maps without $\mathcal{L}_{inter}$ are inconsistent. Pay special attention to the \#2 and \#3 frames.}
	\label{fig:3}
\end{figure*}

%% file: Appendix/fig4.tex
\begin{figure*}[t]
	\centering
	\includegraphics[width=0.65\textwidth]{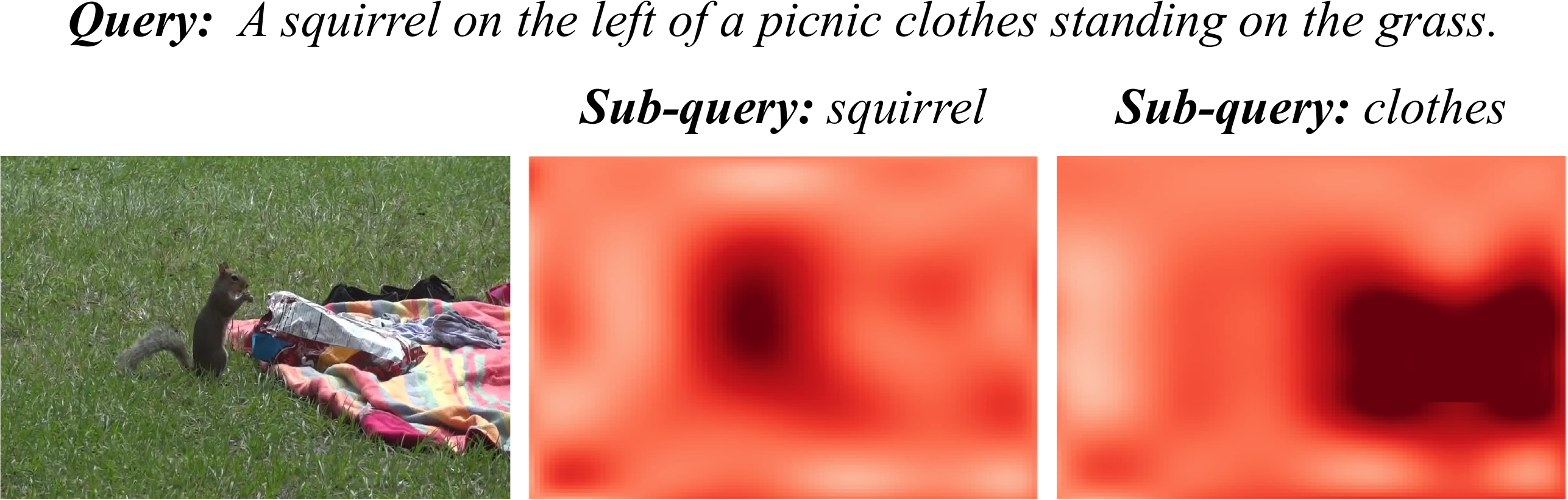}
	\caption{Visualizations of the cross-modal alignment. \emph{Sub-query} is the specific word taken from the full \emph{query}. The response map is generated by computing the cosine similarity between the sub-query and the input image.}
	\label{fig:4}
\end{figure*}

%% file: Appendix/fig5.tex
\begin{figure*}[t]
	\centering
	\includegraphics[width=0.65\textwidth]{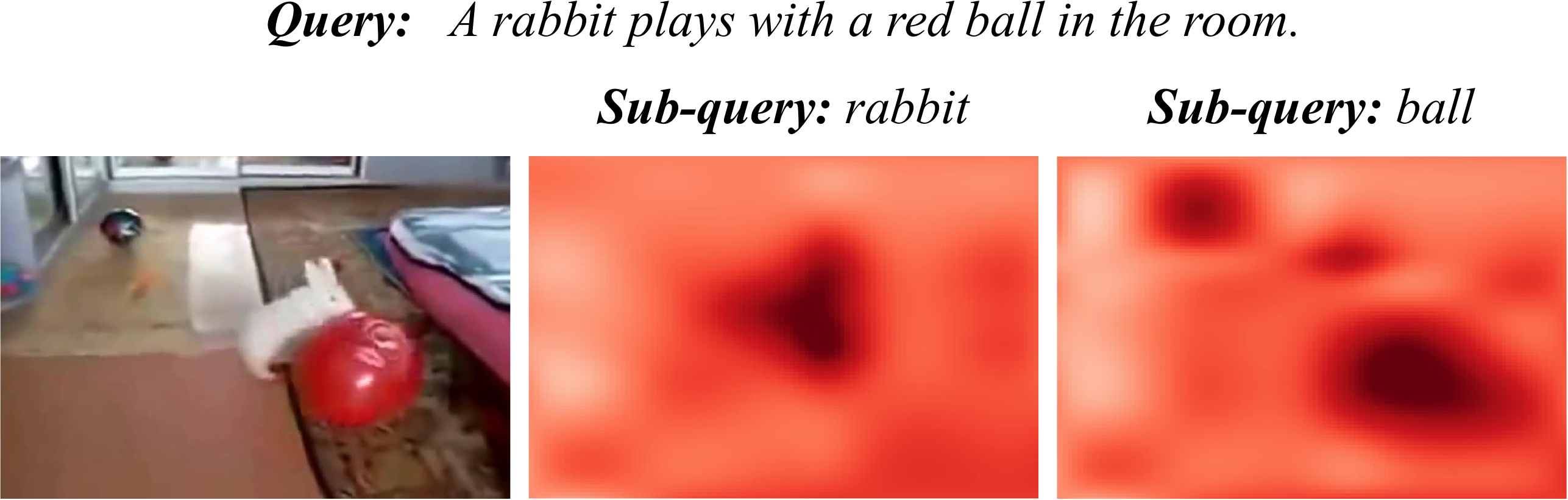}
	\caption{Visualizations of the cross-modal alignment. \emph{Sub-query} is the specific word taken from the full \emph{query}. The response map is generated by computing the cosine similarity between the sub-query and the input image.}
	\label{fig:5}
\end{figure*}